\pdfoutput=1
\documentclass[twoside]{article}

\usepackage[preprint]{aistats2027}

\usepackage[round]{natbib}

\usepackage{amsmath,amssymb,mathtools}
\usepackage{graphicx}
\usepackage{booktabs}
\usepackage{longtable}
\usepackage{placeins}
\usepackage{array}
\usepackage{microtype}
\usepackage[hidelinks]{hyperref}

\graphicspath{{figures/}}

\newcommand{\ec}{e_{c}}
\newcommand{\chis}{\chi_{S}}
\newcommand{\heat}{C_{H}}
\newcommand{\Kvec}{\mathbf{K}}
\newcommand{\Kpvec}{\mathbf{K}'}
\newcommand{\Tmat}{\mathbf{T}}
\newcommand{\Hloss}{H_{\mathrm{loss}}}

\begin{document}

\runningauthor{An Effective Renormalization-Group Description of Grokking}
\runningtitle{An Effective Renormalization-Group Description of Grokking}

\twocolumn[

\aistatstitle{Scaling and Renormalization in Neural Networks: An Effective Renormalization-Group Description of Grokking}

\aistatsauthor{ Min Chen$^{1}$ \And Junqi Wang$^{1}$ \And Zeguan Wu$^{1}$ \And Wenhao Huang$^{2}$ \And Junyu Liu$^{1}$ }

\aistatsaddress{ $^{1}$Department of Computer Science, University of Pittsburgh, Pittsburgh, PA, USA \\ $^{2}$m-a-p.ai } ]
\fancyhead[CO]{}\fancyhead[CE]{}\renewcommand{\headrulewidth}{0pt}

\begin{abstract}
Grokking is a delayed transition from memorization to generalization in deep neural networks (NNs). We examine whether scaling and renormalization group (RG) theories can describe this dynamics, establishing a quantitative physical framework for neural generalization transitions. For modular addition on a two-layer NN, an ensemble of trained models defines an order parameter (the squared weight norm) and fluctuation proxies for the heat capacity and the susceptibility, with training epoch and weight decay serving as thermal and field control coordinates. To bridge neural architectures to real-space RG, we construct an exact coupling matrix via a second-order expansion of the loss with respect to per-neuron norms and use its Fiedler vector to define an effective geometry. This spectral ordering yields a monotonic spatial decay of coupling strength that is absent under random permutations. We then execute real-space RG by block-spin aggregating adjacent neurons and parameterizing the coupling matrix into off-diagonal band strengths. Linearizing this local RG map yields relevant scaling exponents and five derived critical exponents, with direct empirical estimates matching RG predictions within $1$--$10\%$. Overall, our framework provides a consistent effective description of grokking, suggesting the scaling and renormalization group formalism as a tool to investigate NN dynamics.
\end{abstract}

\section{Introduction}
\label{sec:intro}

Grokking refers to the delayed generalization where NNs generalize long after
overfitting their training set~\citep{power2022grokking}. While training accuracy rapidly saturates, test performance remains suppressed before abruptly rising within a narrow epoch window. This sharp transition has inspired thermodynamic and statistical physics interpretations, including effective field theories and phase diagrams \citep{liu2022effective}, weight-norm mechanisms \citep{liu2023omnigrok}, mechanistic progress measures \citep{nanda2023progress}, lazy-to-rich transitions \citep{kumar2024lazy}, first-order mappings in two-layer models \citep{rubin2024firstorder}, and finite-size scaling interpretations \citep{bi2026falsifiable,wang2026dimensional}. Analogous phase-transition phenomena are observed in the emergent abilities of scaling language models \citep{wei2022emergent}, where the perceived sharpness can depend on the choice of metric \citep{schaeffer2023mirage}. Within this broader context, grokking offers a compact, controlled testbed for quantitatively evaluating these statistical physics analogies.

\begin{figure*}[t]
\centering
\includegraphics[width=\textwidth]{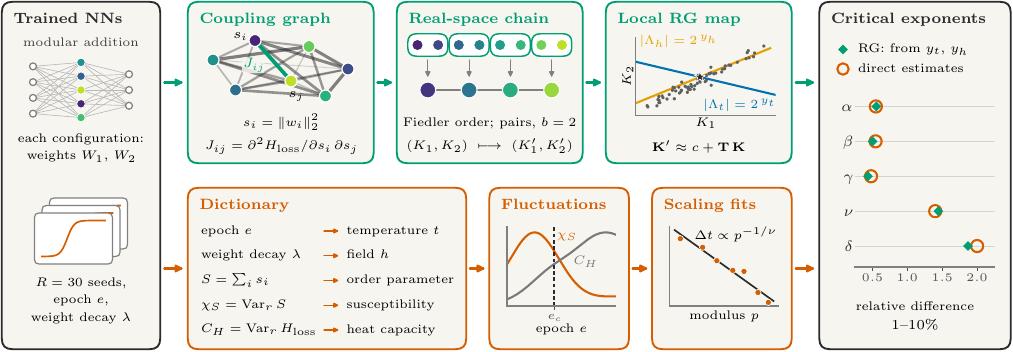}
\caption{Overview. From trained NNs (left), the critical exponents of grokking are obtained via two complementary routes. \emph{RG route} (top): each hidden neuron carries a spin $s_i$, its squared weight norm, while each neuron pair is connected by a coupling $J_{ij}$ derived from the loss Hessian. Ordering the neurons on a chain using the Fiedler vector of this coupling graph enables adjacent neurons to be merged into blocks, where a linear fit of the coarsened couplings yields $y_t$ and $y_h$, from which five critical exponents follow (Sections~\ref{sec:space}--\ref{sec:rg}). \emph{Direct route} (bottom): a physical dictionary uses the training epoch as an effective temperature, weight decay as conjugate field, and ensemble fluctuations as proxies for susceptibility and heat capacity, with power-law fits of these observables yielding the direct exponents (Sections~\ref{sec:setup} and~\ref{sec:direct}). The two independent estimates agree within $1$--$10\%$ (right). }
\label{fig:framework}
\end{figure*}

The renormalization group (RG) provides a quantitative framework for analyzing such transitions. At its core, RG operates via coarse-graining: rescaling a system by a spatial factor $b$ maps its coupling constants to a new set of effective couplings. Near a critical point, the linearized mapping yields eigenvalues $b^{y_i}$, where the two relevant exponents $y_t$ and $y_h$ fully determine all standard critical exponents \citep{wilson1975rg,cardy1996scaling}. Transporting this scaling picture to deep learning promises a unified quantitative theory of grokking, yet applying the RG  to NNs is far from straightforward: First, there is no established correspondence between the quantities of a scaling analysis (e.g., temperature, field, order parameter, susceptibility) and the observables of a training run, so each one has to be identified, and the statistical ensemble is formed by independent training runs instead of by a Gibbs distribution. Second, real-space RG coarse-grains degrees of freedom that are close to each other, yet hidden neurons carry no positions: they are interchangeable under permutation, and the NN defines neither a distance between them nor a lattice on which blocks could be formed.

We make each ingredient operational, as outlined in Figure~\ref{fig:framework}. Our contributions are:
\begin{itemize}
    \item a physical dictionary that maps each thermodynamic quantity, including temperature, field, order parameter, and susceptibility, onto a measurable NN observable. This dictionary is derived from an effective Gibbs canonical ensemble of the loss objective, wherein weight decay acts as the field conjugate to the squared weight norm, cross-run variances serve as proxies for the susceptibility and heat capacity by analogy with equilibrium fluctuation--response relations, and the training epoch serves as an effective inverse temperature (Appendix~\ref{app:gibbs}), with all underlying assumptions explicitly detailed.
    \item a construction of a real space for hidden neurons, which the NN does not provide: treating the per-neuron norm coordinates as spins and the training loss as the energy, the coupling $J_{ij}$ between two neurons is the second derivative of the loss with respect to their norm coordinates, computed exactly by rescaling the activation of each neuron; the matrix $\lvert J_{ij}\rvert$ is then read as a weighted graph, and ordering the neurons along the second eigenvector of its Laplacian (the Fiedler vector) places them on a one-dimensional chain with a distance $r_{ij}$, along which the mean coupling decays monotonically, whereas a random ordering gives a flat profile (Section~\ref{sec:space} and Appendix~\ref{app:realspace}).
    \item a local RG map for the NN: adjacent neurons in the constructed real space are merged into block spins ($b=2$), the coupling matrix is reduced to two coordinates (the mean nearest- and next-nearest-neighbour couplings), the map from the original to the blocked coordinates is fitted by a weighted regression over NN configurations around the grokking center, and its eigenvalues give the scaling exponents $y_t$ and $y_h$ and, through the standard scaling relations, five critical exponents (Section~\ref{sec:rg}). Direct estimates of the exponents (measured through the dictionary of the first contribution) agree with the RG values within 1--10\%.
\end{itemize}

\section{Related Work}
\label{sec:related}

The original grokking experiments identified delayed generalization on small algorithmic datasets under prolonged optimization with regularization \citep{power2022grokking}. The effect is not specific to these datasets or to NNs: it appears on images, text, and molecules when the initial weight norm is large \citep{liu2023omnigrok}, in group multiplication \citep{stander2023grokking} and in learning local rules \citep{vzunkovivc2024grokking}, and in non-neural feature-learning models \citep{mallinar2024emergence}. Most explanations describe a competition between a memorizing and a generalizing solution: an effective theory of representation learning with phase diagrams over data and optimization parameters \citep{liu2022effective}, Fourier circuits and continuous progress measures for modular addition \citep{nanda2023progress}, and the higher efficiency of the generalizing circuit \citep{varma2023explaining}; the Fourier structure of modular addition has also been analysed directly \citep{gromov2023grokking,mohamadi2024you}. Other accounts place the delay in the optimization dynamics, as a transition from lazy to rich training \citep{kumar2024lazy} or a change of implicit bias between early and late training \citep{lyu2024dichotomy}, reproduce it in solvable linear models \citep{levi2024grokking,beck2024grokking}, or tie it to the edge of numerical stability \citep{prieto2025grokking}; the delay can also be shortened by transferring embeddings from a weaker model \citep{xu2025let}. \citet{huang2024unified} relate grokking to double descent and to emergent abilities.

Abrupt changes with training or scale are not limited to grokking. Large language models show emergent abilities that appear suddenly as the model grows \citep{wei2022emergent}; whether these are sharp transitions or an effect of discontinuous evaluation metrics is debated \citep{schaeffer2023mirage,du2024understanding,DBLP:conf/iclr/HuL0ZHZL0OZ0024}, and \citet{lubana2025a} model emergence as a percolation transition in transformers trained on a formal language. Power-law scaling of performance with model and data size has likewise been explained with methods from theoretical physics, including large-$N$ field theory \citep{doi:10.1073/pnas.2311878121,zhang2025neural}.

Several works describe grokking in the language of phase transitions. \citet{rubin2024firstorder} gave a first-order phase-transition account for a class of two-layer NNs, \citet{vzunkovivc2024grokking} analysed grokking phase transitions in learning local rules with gradient descent, \citet{pomarico2025grokking} described grokking as an entanglement transition in tensor-network models of learning, and recent works study finite-size signatures \citep{bi2026falsifiable,wang2026dimensional}. 

Existing studies typically frame grokking as a phase transition through isolated observables, yet they lack a derivation of scaling behavior rooted in a physical coarse-graining of the NN's internal representations. The renormalization group achieves this in statistical mechanics by systematically coarsening microscopic degrees of freedom and identifying the relevant scaling directions near a critical point \citep{wilson1975rg,cardy1996scaling}, whereas prior connections between RG and deep learning have remained predominantly conceptual \citep{mehta2014mapping}. To bridge this gap, we implement an empirical RG transformation on the hidden neurons. Specifically, a Hessian-based coupling induces a functional neuron graph, the second eigenvector of its graph Laplacian \citep{fiedler1973algebraic} arranges the neurons along a one-dimensional chain, and neighboring neurons are merged via spatial blocking.

\section{Methods}\label{sec:methods}
\subsection{Setup and Observables}
\label{sec:setup}

\paragraph{Task, model, and ensemble.}

We consider modular addition, $y = (x_1 + x_2) \bmod p$, with $p = 97$ following the standard setup of \citet{power2022grokking}. Each operand is represented as a one-hot vector, resulting in a $2p = 194$-dimensional input. Following \citet{gromov2023grokking}, we employ a bias-free two-layer perceptron with quadratic activation,
\begin{equation}
    f_\theta(x)=W_2\,(W_1x)^{\odot 2},
\end{equation}
with $W_1\in\mathbb{R}^{N\times 2p}$, $W_2\in\mathbb{R}^{p\times N}$, and $N=500$ hidden units. The model is trained on a $50\%$ random split of all $p^2$ ordered pairs using full-batch mean squared error (MSE) loss on one-hot targets. Optimization is performed using AdamW \citep{loshchilov2019adamw} with a learning rate of $10^{-3}$, weight decay $\lambda = 15$, and momentum parameters $(\beta_1, \beta_2) = (0.9, 0.98)$. The experiments are run on a single NVIDIA RTX 4090.

Our primary benchmark consists of an ensemble of $R = 30$ independent runs. For run $r$, seed $r$ controls both initialization and the dataset split, varying both sources of stochasticity simultaneously. All 30 runs achieve $99\%$ test accuracy between epochs 170 and 182. We refer to this collection as an \emph{empirical-seed ensemble}. Crucially, it differs from a traditional Gibbs ensemble: its fluctuations reflect a joint combination of optimization dynamics and data-sampling variability; since each split is held fixed during training, the latter acts as quenched disorder, and averages over runs are quenched averages, which in principle differ from thermal averages at fixed disorder. Consequently, all thermodynamic terms introduced hereafter should be interpreted as effective descriptions.

\paragraph{Order parameter, fluctuations, and control coordinates.}

For each hidden unit $i$, we concatenate its incoming and outgoing weight vectors and compute its squared $L_2$ norm,
\begin{equation}
\begin{aligned}
    w_i &= \operatorname{concat}(W_{1,i:}, W_{2,:i}), \\
    s_i &= \|w_i\|_2^2, \qquad S = \sum_{i=1}^{N} s_i.
\end{aligned}
\label{eq:order}
\end{equation}
The total squared norm $S$ serves as the order parameter upon which weight decay acts, given that the regularized loss takes the form $H_{\mathrm{wd}} = \Hloss + (\lambda / 2) S$, establishing $\lambda$ as the natural conjugate field (analogous to the external magnetic field $h$ coupling to the magnetization $M$ in the Ising model). The Ising model serves here as a standard point of comparison; unlike its magnetization, however, $S$ is nonnegative and remains nonzero on both sides of the transition At epoch $e$, we record the cross-seed variance of $S$ and the training loss $\Hloss$:
\begin{equation}
    \chis(e) = \operatorname{Var}_r[S_r(e)], \qquad \heat(e) = \operatorname{Var}_r[\Hloss{}_{,r}(e)],
    \label{eq:fluct}
\end{equation}
which serve as proxies for the susceptibility and heat capacity by analogy with the equilibrium fluctuation forms, $\chi \propto \operatorname{Var}(M)$ and $C \propto \operatorname{Var}(E)$. Note that we omit the variance of $H_{\mathrm{wd}}$ when defining the heat capacity to prevent the terms $(\lambda^2 / 4) \operatorname{Var}(S)$ and $\lambda\operatorname{Cov}(\Hloss,S)$ from mixing susceptibility contributions into it.

\begin{figure}[t]
    \centering
    \includegraphics[width=\columnwidth]{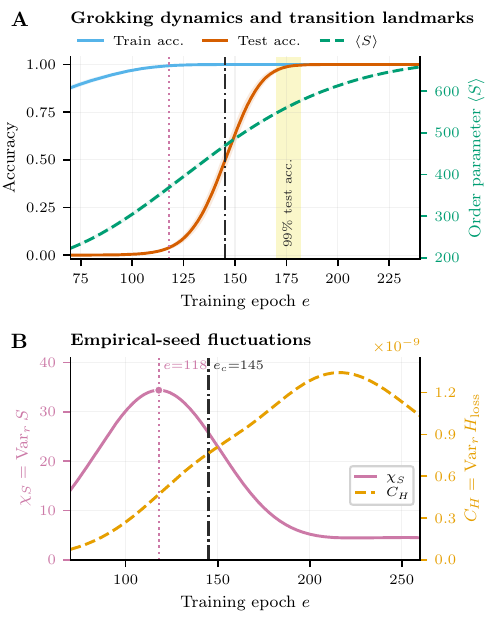}
    \caption{Ensemble dynamics. (A) Train and test accuracy and the order parameter $\langle S\rangle$ over the 30 runs. (B) Susceptibility $\chis=\operatorname{Var}_r S$ and heat-capacity proxy $\heat=\operatorname{Var}_r\Hloss$. The dotted line marks the $\chis$ peak at $e=118$; the dash-dotted line marks $\ec=145$.}
    \label{fig:dynamics}
\end{figure}

Two key landmarks characterize the transition (Figure~\ref{fig:dynamics}): the susceptibility $\chis$ peaks at $e = 118$, while the ensemble-mean test accuracy exhibits its maximum slope at $e_c = 145$. Because grokking is inherently a generalization phenomenon, we align our theoretical analysis at $e_c$ and define the reduced temperature $t$ and field $h$ as
\begin{equation}
    t = \frac{e - e_c}{e_c}, \qquad h = -\frac{\lambda - \lambda_c}{2}, \qquad \lambda_c = 15.
    \label{eq:th}
\end{equation}
The sign of $h$ naturally arises from rewriting $H_{\mathrm{wd}} = [\Hloss + (\lambda_c / 2) S] - h S$, ensuring that $h$ couples to $S$ under the standard Ising field convention. Assuming an effective inverse temperature that grows linearly with training, $\beta_{\mathrm{eff}} \propto e$, the thermodynamic reduced temperature becomes $t_T = e_c / e - 1 = -t / (1 + t) \approx -t$ to leading order. Consequently, $t < 0$ corresponds to the disordered phase; all scaling exponents reported below are defined with respect to $|t|$. Appendix~\ref{app:gibbs} details the motivation for this physical mapping and the corresponding Gibbs representation.

\paragraph{Dictionary.}

Table~\ref{tab:dictionary} in Appendix~\ref{app:gibbs} summarizes each observable in our scaling analysis, alongside its standard statistical-physics definition \citep{cardy1996scaling} and its operational counterpart within the NN. Notably, two key components require explicit mathematical construction instead of direct measurement: the effective lattice distance and the block-spin transformation. Section~\ref{sec:space} constructs the spatial distance metric, while Section~\ref{sec:rg} introduces the coarse-graining transformation.

\subsection{Constructing the Real Space}
\label{sec:space}

\paragraph{From norm coordinates to a coupling matrix.}

In our dictionary, the per-neuron norm coordinates $s_i$ in Eq.~\eqref{eq:order} play the role of spin variables, while the training loss $\Hloss$ represents the total energy. Expanding $\Hloss$ to second order around a trained state $s^0$,
\begin{equation}
    \Hloss(s) \approx \Hloss(s^0) + \sum_i g_i \delta s_i + \frac{1}{2} \sum_{i,j} J_{ij} \delta s_i \delta s_j,
\end{equation}
defines the coupling matrix via the pairwise interaction coefficients, $J_{ij} = \frac{\partial^2 \Hloss}{\partial s_i \partial s_j}$. To perturb a single coordinate $s_i$, we scale the activations of neuron $i$ by a scalar factor $a_i$, so that the norm coordinate transforms as $s_i(a_i) = a_i^2 s_i$, where $s_i$ denotes the unperturbed baseline value (at $a_i = 1$). For the mean-squared-error loss, with hidden activations $\sigma_{ni}=[(W_1x_n)_i]^2$ and $n_y$ scalar targets, the off-diagonal entries have the closed form
\begin{equation}
    J_{ij}=\frac{(\sigma^{\top}\sigma)_{ij}\,(W_2^{\top}W_2)_{ij}}{2\,n_y\,s_i\,s_j}\qquad(i\neq j).
    \label{eq:J}
\end{equation}
Intuitively, two neurons are strongly coupled when both their activation patterns across the training set and their output weight vectors overlap significantly. The diagonal elements $J_{ii}$ contain an extra term due to the quadratic dependence of $s_i(a_i)$ on the scaling factor $a_i$, and are omitted in all downstream analyses. To avoid numerical instabilities, dead neurons satisfying $s_i < 10^{-8} \operatorname{median}_j s_j$ are masked. Notably, no neurons are masked at $e_c$.

\paragraph{Effective lattice coordinates via spectral ordering.}
\label{sec:space_ordering}

Unlike standard physical spin systems defined on a fixed geometric lattice, hidden neurons in an NN lack a native spatial ordering. To construct a meaningful notion of distance, we treat the absolute couplings $A_{ij} = |J_{ij}|$ (with $A_{ii} = 0$) as the adjacency matrix of a similarity graph among neurons. We then compute the unnormalized graph Laplacian $\mathcal{L} = D - A$, where $D_{ii} = \sum_j A_{ij}$ denotes the degree matrix.  To embed the neurons into an effective one-dimensional lattice, we utilize the Fiedler vector $v_2$, which is the eigenvector corresponding to the algebraic connectivity (the second-smallest eigenvalue) of $\mathcal{L}$ \citep{fiedler1973algebraic}. By minimizing $\sum_{i<j} A_{ij}(v_i - v_j)^2$ subject to $\|v\|_2 = 1$ and $v \perp \mathbf{1}$, the Fiedler vector maps strongly coupled neurons closer to one another along a continuous line. Finally, we assign discrete spatial positions $\pi(i) \in \{1, \dots, N\}$ based on the ranked order of the components $v_{2,i}$, defining the pairwise lattice distance as $r_{ij} = |\pi(i) - \pi(j)|$. This spectral ordering is recomputed dynamically for every NN configuration (each run at each epoch) to reflect the evolving coupling structure.

\paragraph{Emergence of spatial locality.}
\label{sec:space_validation}

\begin{figure*}[t]
    \centering
    \includegraphics[height=0.245\textwidth]{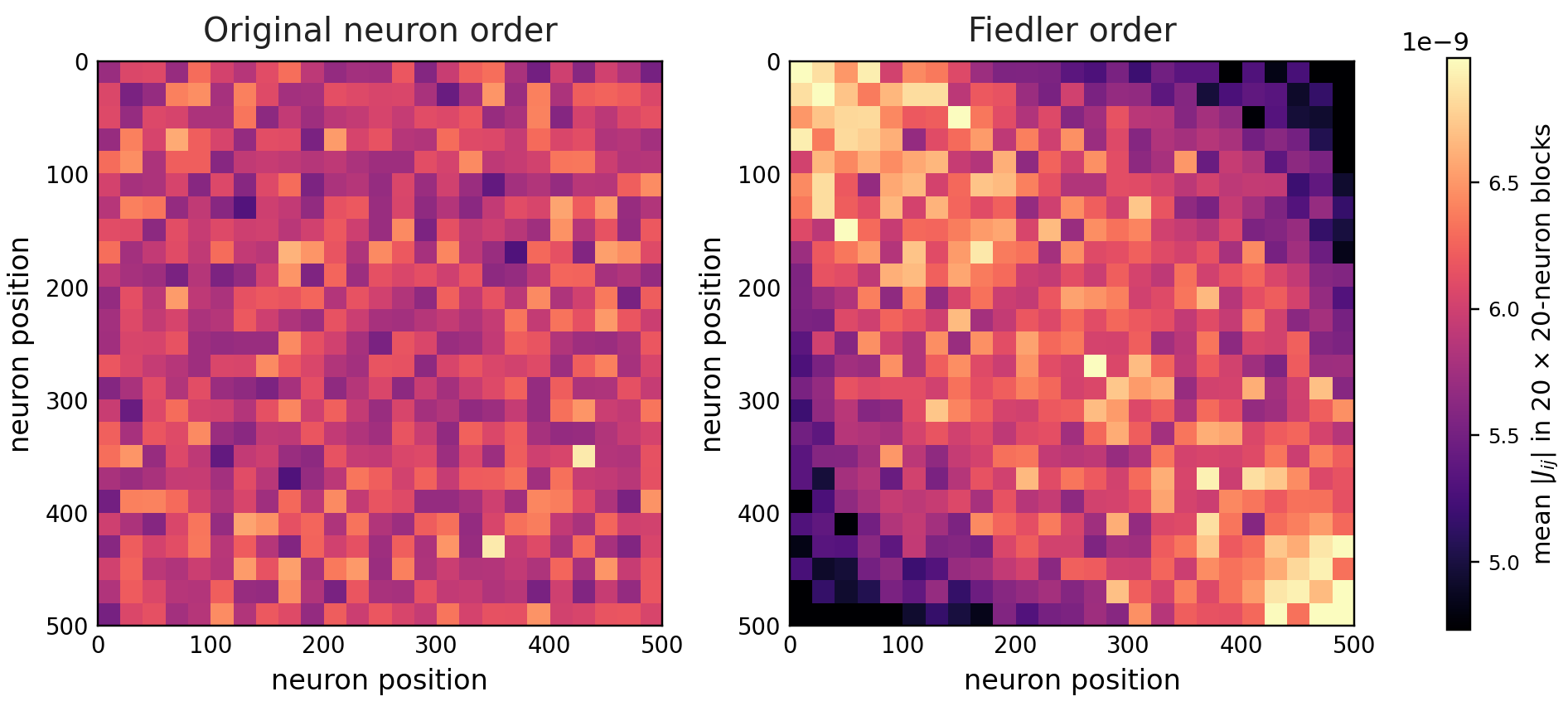}\hfill
    \includegraphics[height=0.245\textwidth]{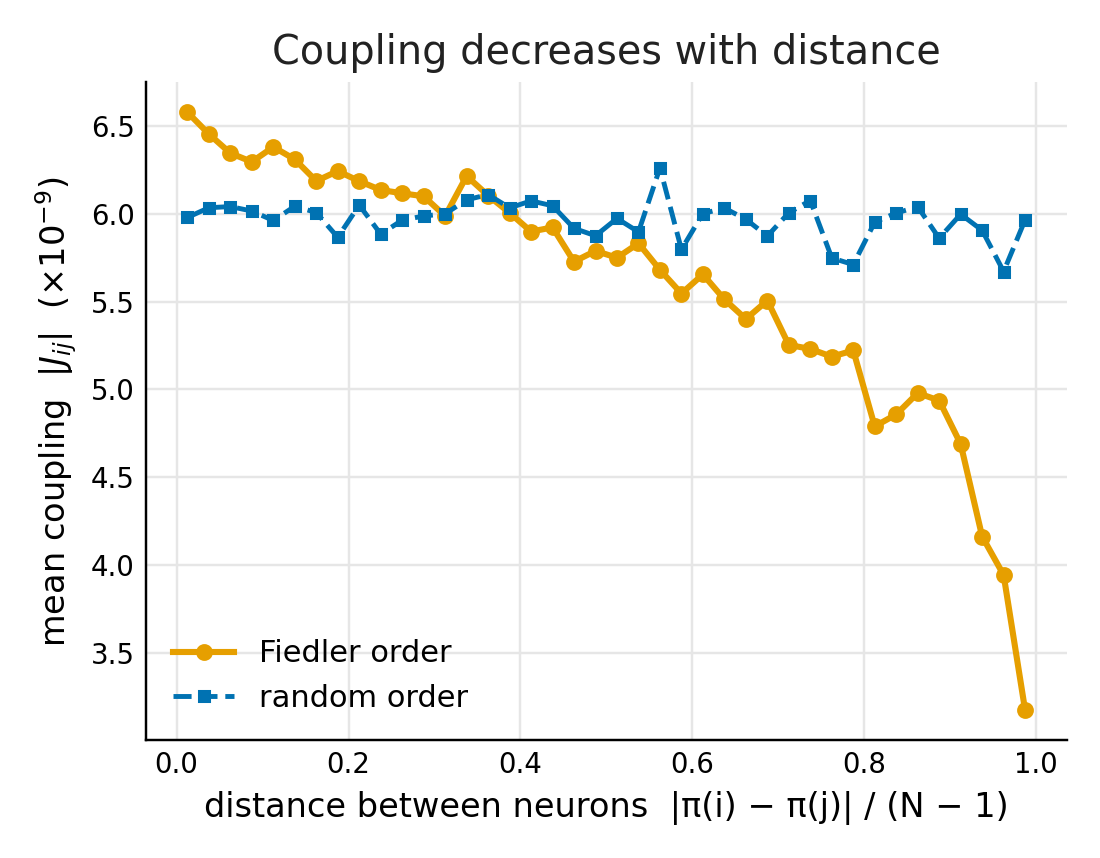}
    \caption{Validation of the constructed real space at $\ec=145$. Left pair: $|J_{ij}|$ averaged over $20\times20$-neuron blocks, in the original neuron order and after Fiedler ordering, on a common colour scale. Right: mean $|J_{ij}|$ between neuron pairs at a given normalized Fiedler distance, compared with a random ordering of the same matrix.}
    \label{fig:space}
\end{figure*}

Figure~\ref{fig:space} shows that our spectral ordering construct generates an effective geometry in which coupling strength decays with spatial distance. When coarse-graining the $500 \times 500$ coupling matrix $|J_{ij}|$ at $e_c$ into $20 \times 20$-neuron blocks, the original unordered index reveals no discernible spatial structure. In contrast, the Fiedler-ordered matrix exhibits a prominent diagonal band structure, revealing a near-diagonal concentration of coupling strength (Figure~\ref{fig:space}).  Quantitatively, the mean coupling strength decreases monotonically as a function of the normalized distance $r_{ij} / (N - 1)$. Specifically, the nearest $2.5\%$ of neuron pairs exhibit a mean coupling strength $2.1\times$ larger than that of the farthest $2.5\%$, whereas a baseline random ordering yields a completely flat distribution with no spatial dependence. Additional details of the construction and distance-profile analysis are provided in Appendix~\ref{app:realspace}.

\subsection{Renormalization Group}
\label{sec:rg}

\paragraph{Blocking and two coupling coordinates.}

A standard real-space renormalization group (RG) step consists of three core components \citep{wilson1975rg,cardy1996scaling}: 
(i)~integrating out short-distance degrees of freedom via a projection or blocking map $\mathcal{B}$,
\begin{equation}
    e^{-\mathcal{A}'[\phi']} = \int d\phi \, \delta(\phi' - \mathcal{B}\phi) \, e^{-\mathcal{A}[\phi]}, \quad \mathcal{A} = H/k_BT;
\end{equation}
(ii)~rescaling coarse-grained variables and spatial coordinates; and 
(iii)~tracking the induced flow of coupling constants $\Kpvec = R_b(\Kvec)$.
Our construction adapts two of these steps directly to the neural setting: (i)~coarse-graining is executed via an explicit block-spin aggregation rule and (ii)~the resulting RG flow is mapped by tracking two primary induced coupling parameters. Appendix~\ref{app:rgbackground} reviews the underlying theory, from block spins to exponent formulas, for readers unfamiliar with the real-space RG formalism.

The block-spin transformation groups adjacent neuron pairs in Fiedler order into blocks $B_k = \{\pi^{-1}(2k-1), \pi^{-1}(2k)\}$ with a spatial scale factor $b = 2$. The coarse-grained spin variables and microscopic couplings are aggregated as
\begin{equation}
    s'_B = \frac{1}{b}\sum_{i \in B} s_i, \qquad J'_{BB'} = \sum_{i \in B} \sum_{j \in B'} J_{ij},
    \label{eq:block}
\end{equation}
which maps $500$ individual neurons to $250$ coarse-grained blocks. We take the block mean as block variable, so that $J'_{BB'}$ is the coupling of the quadratic expansion above restricted to block-uniform perturbations, $\delta s_i = \delta s'_B$ for $i \in B$. To represent the full $500 \times 500$ coupling matrix compactly, we define two scalar parameters, $K_1$ and $K_2$, corresponding to the mean absolute coupling along the first and second off-diagonal bands in Fiedler order:
\begin{equation}
\begin{aligned}
    K_1 &= \frac{1}{N-1} \sum_{i=1}^{N-1} \big| J_{\pi^{-1}(i)\, \pi^{-1}(i+1)} \big|, \\
    K_2 &= \frac{1}{N-2} \sum_{i=1}^{N-2} \big| J_{\pi^{-1}(i)\, \pi^{-1}(i+2)} \big|.
\end{aligned}
    \label{eq:K}
\end{equation}
Applying the same definition to the coarse-grained coupling $J'$ yields $\Kpvec = (K'_1, K'_2)$. Truncating the coupling space to these two coordinates $\Kvec = (K_1, K_2)$ provides a low-dimensional manifold, enabling us to estimate the linearized RG transformation $\Kvec \to \Kpvec$ even from a finite ensemble of trained models.

\paragraph{The local RG map.}

Since the mapping $\Kvec \mapsto \Kpvec$ lacks a closed-form analytical expression, we estimate its local linear dynamics empirically. By training an ensemble of models across $13$ epochs $e \in \{115, 120, \dots, 175\}$ and $5$ weight decay values $\lambda \in \{13, \dots, 17\}$, we obtain $65$ empirical pairs $(\Kvec_n, \Kpvec_n)$. To extract the derivative (Jacobian) of the RG map, we perform a local weighted least-squares regression:
\begin{equation}
    \min_{c, \Tmat} \sum_{n=1}^{65} w_n \big\lVert \Kpvec_n - c - \Tmat \Kvec_n \big\rVert_2^2,
    \label{eq:fit}
\end{equation}
where $w_n$ is a two-dimensional Gaussian window function centered at $(\ec, \lambda_c)$ in the $(e, \lambda)$ parameter space (Appendix~\ref{app:fit}). The fitted affine map $\Kpvec \approx c + \Tmat\Kvec$ defines the Jacobian $\Tmat$ of the coarse-graining transformation around the grokking center. The fitted transformation matrix is
\begin{equation}
    \Tmat = \begin{pmatrix} -0.4458 & 2.7169 \\ 0.8685 & 0.4008 \end{pmatrix}.
    \label{eq:T}
\end{equation}

\paragraph{Scaling fields.}
The local linear RG dynamics $\Tmat$ yields two real eigenvalues, $\Lambda_+ = +1.5708$ and $\Lambda_- = -1.6159$, with normalized right eigenvectors $v_+ = (0.803, 0.596)^\top$ and $v_- = (0.918, -0.396)^\top$. Geometrically, these right eigenvectors $v_\pm$ define the invariant axes in parameter space along which coupling perturbations stretch or shrink without spatial rotation. However, because $\Tmat$ is non-symmetric ($\Tmat \neq \Tmat^\top$), these right eigenvectors are non-orthogonal, meaning that projections onto $v_\pm$ would mix under RG transformations. To resolve this, we leverage the biorthogonality between left and right eigenvectors ($\ell_a^\top v_{a'} = \delta_{aa'}$): projecting perturbations onto the left eigenvectors via $u_a = \ell_a^{\top} \delta\Kvec$ eliminates cross-mode interference, yielding effective local scaling directions that decouple independently as $u_a' = \Lambda_a u_a$.

Unlike standard physical spin systems, where internal discrete symmetries cleanly decouple thermal and magnetic modes, the parameter spaces of NNs lack such \textit{a priori} symmetries. We therefore assign physical identities to these modes by quantifying their alignment with physical control coordinates in parameter space. Specifically, we evaluate the tangent vectors $\partial\Kvec/\partial e$ and $\partial\Kvec/\partial\lambda$ along the parameter cloud at the grokking center, corresponding to perturbations driven by epoch ($e$, thermal) and weight decay ($\lambda$, magnetic field), respectively. The weight-decay direction aligns most strongly with $v_+$, so we label $v_+$ as the magnetic direction and $v_-$ as the thermal direction, giving $\Lambda_h = \Lambda_+$ and $\Lambda_t = \Lambda_-$ (see Figure~\ref{fig:alignment} in Appendix~\ref{app:fit}).

A single blocking step with scale factor $b=2$ rescales perturbations along the principal scaling axes by $\Lambda = b^y$, defining the effective local scaling exponents:
\begin{equation}
    y_t = \log_2 |\Lambda_t| = 0.6923, \qquad y_h = \log_2 |\Lambda_h| = 0.6515.
    \label{eq:y}
\end{equation}
Because both eigenvalues satisfy $|\Lambda| > 1$ (hence $y_t, y_h > 0$), both scaling fields are relevant under the RG flow. Two relevant directions, i.e.\ a critical manifold of codimension two, are consistent with an effective critical description controlled simultaneously by thermal (epoch) and field (weight decay) coordinates. The negative sign of the thermal eigenvalue ($\Lambda_t = -1.6159$) reveals an alternating sign dynamics: the thermal perturbations invert their orientation across successive spatial blockings while expanding in magnitude. We attribute this to the structure of the truncated map, whose off-diagonal entries dominate ($\operatorname{tr}\Tmat = -0.045$, so that $\Lambda_\pm \approx \pm|\det\Tmat|^{1/2}$), rather than to a physical alternation. To ensure a well-defined physical scaling exponent, we use the two-step map ($b=4$) with positive eigenvalue $\Lambda_t^2 > 1$, which yields the identical exponent via $y_t = \frac{1}{2}\log_2(\Lambda_t^2) = \log_2|\Lambda_t|$.

\paragraph{Critical exponents.}

Under the additional scaling hypothesis that these effective local directions obey the standard homogeneity form, the singular part of the free-energy density $f_s(t,h)$ is a generalized homogeneous function of the scaling fields \citep{cardy1996scaling}:
\begin{equation}
    f_s(t,h) = b^{-d} \, f_s\big(b^{y_t}t,\, b^{y_h}h\big),
    \label{eq:homog}
\end{equation}
where $d$ represents the spatial dimension of the underlying lattice. Since the Fiedler spectral ordering maps the NN onto an effective one-dimensional chain, we set the effective dimension $d=1$. Setting the scale factor to $b = |t|^{-1/y_t}$ and differentiating equation~\eqref{eq:homog} yields the canonical scaling relations for the critical exponents (See Appendix~\ref{app:exponents} for detailed derivations):
\begin{equation}
\begin{aligned}
    \alpha &= 2 - \frac{d}{y_t}, & \beta &= \frac{d - y_h}{y_t}, & \gamma &= \frac{2y_h - d}{y_t},\\
    \nu &= \frac{1}{y_t}, & \delta &= \frac{y_h}{d - y_h},
\end{aligned}
\label{eq:exponents}
\end{equation}
where $\nu$ follows from the correlation length rescaling $\xi' = \xi/b$. Substituting the RG scaling exponents from equation~\eqref{eq:y} gives:
\begin{equation}
\begin{aligned}
    (\alpha,\,\beta,\,\gamma) &= (0.556,\, 0.503,\, 0.438),\\
    (\nu,\,\delta) &= (1.444,\, 1.870).
\end{aligned}
    \label{eq:rgvalues}
\end{equation}
By construction, standard scaling and hyperscaling identities (e.g.,  Rushbrooke ($\alpha + 2\beta + \gamma = 2$), Widom ($\gamma = \beta(\delta - 1)$), and Josephson ($d\nu = 2 - \alpha$)) are identically satisfied. Appendix~\ref{app:exponents} details the complete derivation of each scaling formula from the thermodynamic free energy and spatial correlation functions.

\begin{figure}[t]
    \centering
    \includegraphics[width=\columnwidth]{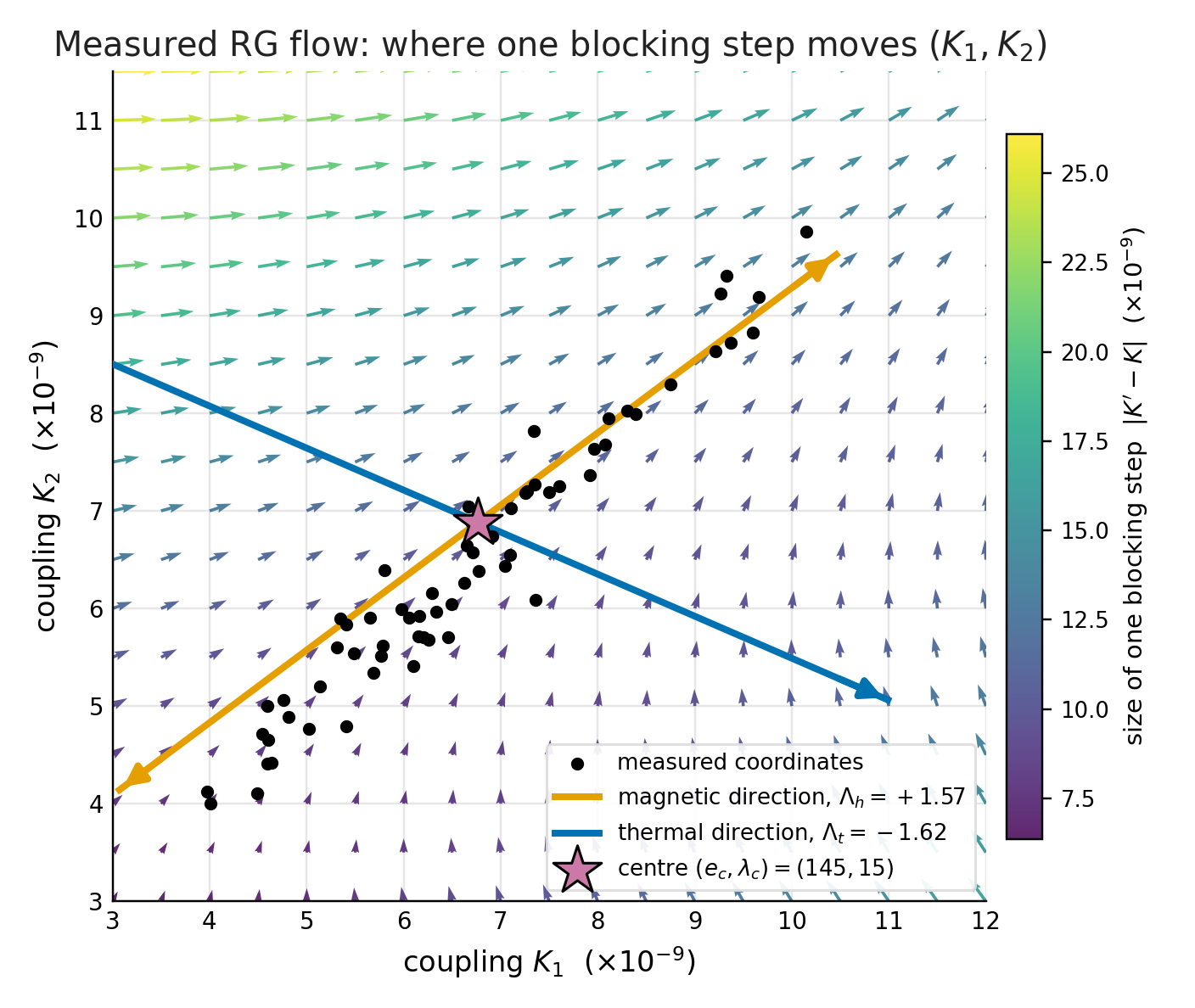}
    \caption{The measured local RG flow. Colour gives the size of the step. Black dots are the 65 measured coordinates. The orange and blue lines are the magnetic and thermal eigendirections of $\Tmat$; both point away from the center because $|\Lambda|>1$. The star is the grokking center $(\ec,\lambda_c)=(145,15)$. The regression that produced $\Tmat$ is shown in Figure~\ref{fig:localfit} (Appendix~\ref{app:fit}).}
    \label{fig:flow}
\end{figure}

Figure~\ref{fig:flow} illustrates the RG flow. Each point of the plane is a pair of couplings $(K_1, K_2)$, and each arrow shows where one blocking step moves that point. The black dots are the 65 measured coordinates. The matrix $\Tmat$ gives its two eigendirections drawn as the orange and blue lines through the grokking center. Two points that differ along the orange (magnetic) line are $1.57$ times farther apart after each step. Two points that differ along the blue (thermal) line are $1.62$ times farther apart after each step, and their order along the line is reversed, which is the meaning of the negative sign of $\Lambda_t$. A difference in any other direction is a combination of these two. Differences therefore grow in every direction and shrink in none. In the language of the RG, both directions are relevant: within this local effective-RG description, remaining near the grokking center requires tuning both control parameters, here the epoch and the weight decay. The two growth factors are the numbers taken from this figure: their base-2 logarithms are the exponents $y_h$ and $y_t$ of equation~\eqref{eq:y}, which determine all five critical exponents. The measured coordinates lie along the magnetic line, and the thermal line crosses them.

\section{Direct Measurements and Comparison}
\label{sec:direct}

\begin{table*}[t]
    \centering
    \small
    \caption{Critical exponents from direct estimates and from the local RG. Relative difference is $|\mathrm{direct}-\mathrm{RG}|/\mathrm{RG}$. }
    \label{tab:exponents}
    \begin{tabular}{clp{6.4cm}ccc}
        \toprule
        Exp. & Definition & How the direct value is obtained & Direct & RG & Rel. diff. \\
        \midrule
        $\alpha$ & $C\propto|t|^{-\alpha}$ & $\heat=\operatorname{Var}_r\Hloss$, $t<0$, $e\in[107,133]$ & 0.549 & 0.556 & 1.2\% \\
        $\beta$ & $m\propto|t|^{\beta}$ & $|\langle S\rangle_r-S_c|$, $t>0$, $e\in[180,256]$ & 0.545 & 0.503 & 8.3\% \\
        $\gamma$ & $\chi\propto|t|^{-\gamma}$ & $\chis=\operatorname{Var}_r S$, $t>0$, $e\in[153,171]$ & 0.480 & 0.438 & 9.7\% \\
        $\nu$ & $\xi\propto|t|^{-\nu}$ & $\operatorname{std}_r e_{50}/\langle e_{50}\rangle_r\propto p^{-1/\nu}$, seven moduli & 1.400 & 1.444 & 3.1\% \\
        $\delta$ & $m\propto|h|^{1/\delta}$, $t=0$ & $|\langle S/N \rangle_h - \langle S/N \rangle_{h=0}|\propto|h|^{1/\delta}$ at $\ec$  & 2.001 & 1.870 & 7.0\% \\
        \bottomrule
    \end{tabular}
\end{table*}

\begin{figure*}[t]
    \centering
    \includegraphics[width=0.98\textwidth]{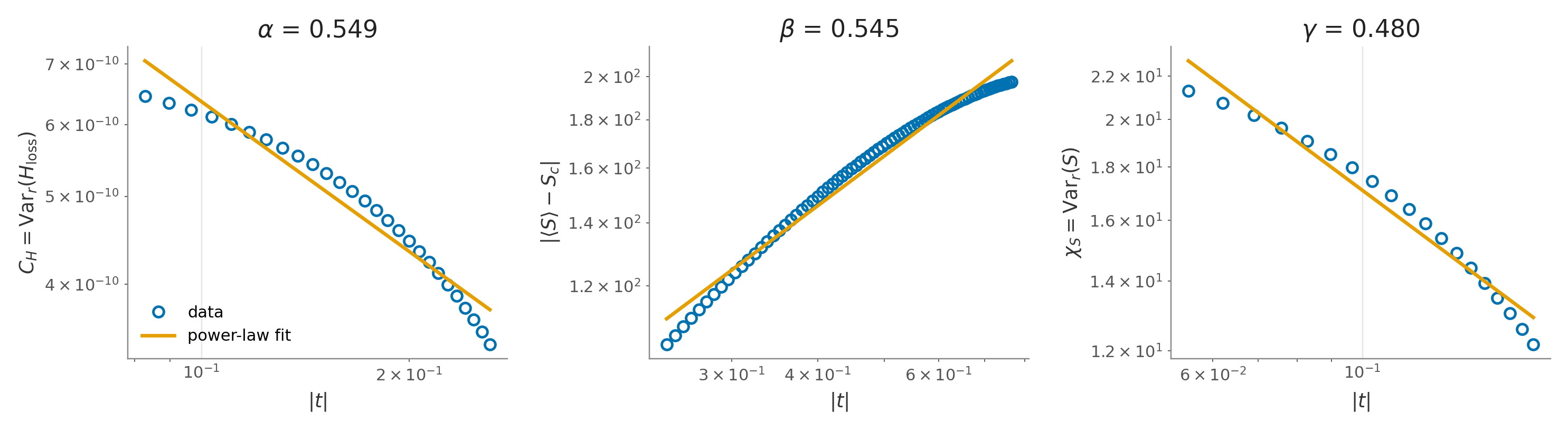}\\[4pt]
    \includegraphics[height=0.23\textwidth]{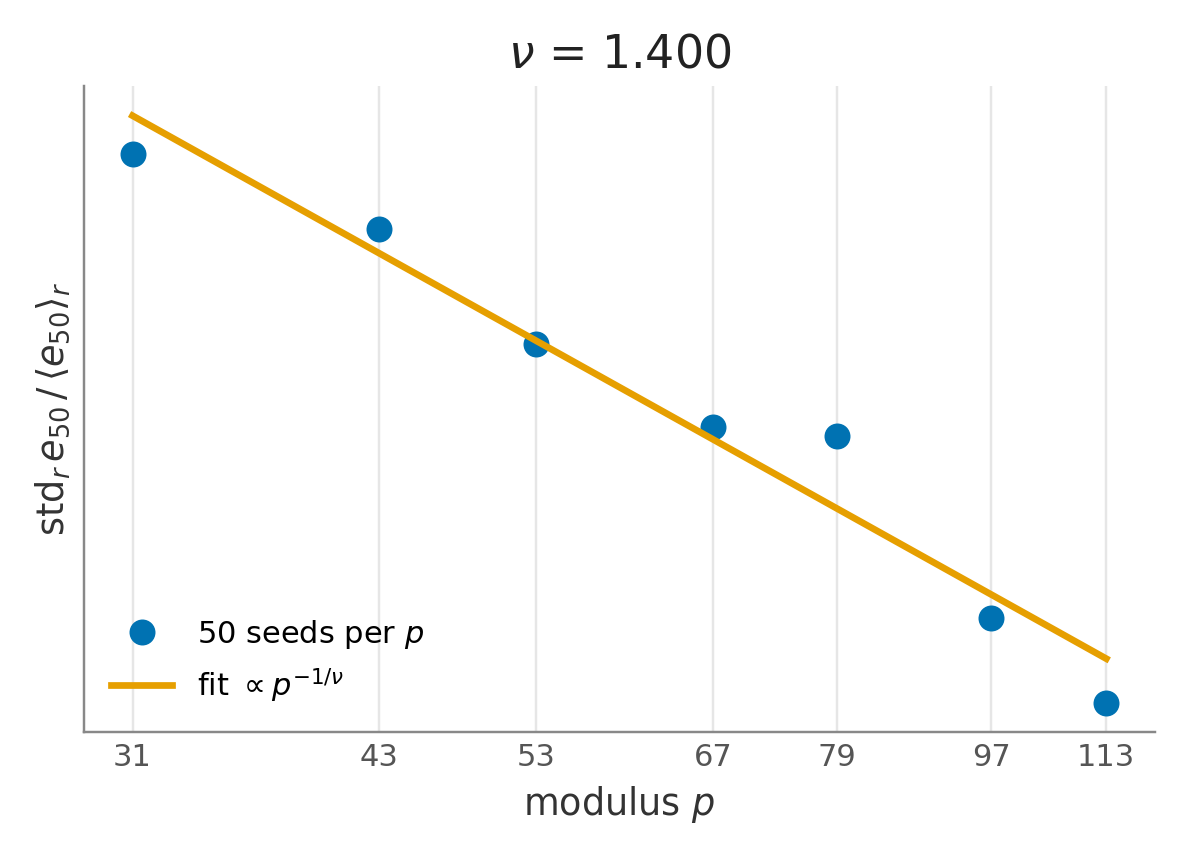}\hspace{0.04\textwidth}
    \includegraphics[height=0.23\textwidth]{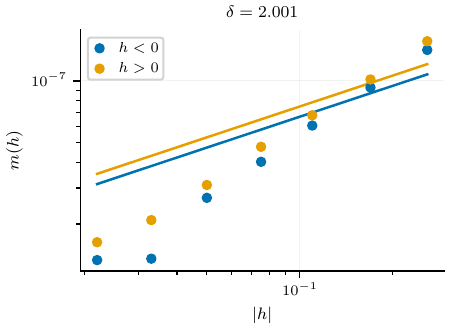}
    \caption{Direct fits for the five critical exponents: $\alpha$, $\beta$, and $\gamma$ (top), $\nu$ (bottom left), and $\delta$ (bottom right).}
    \label{fig:fits}
\end{figure*}

Table~\ref{tab:exponents} contrasts the exponent values predicted by the RG analysis against direct estimates from the observables defined in Section~\ref{sec:setup}. Here, $t=(e-e_c)/e_c$ represents the reduced control variable with $e_c=145$, while $\langle\cdot\rangle_r$ and $\operatorname{Var}_r$ indicate the sample mean and variance across independent runs. Each row matches a thermodynamic scaling relation with its equivalent NN observable from Table~\ref{tab:dictionary}.
The first three exponents are determined from separate log--log least-squares fits against $|t|$ using the main ensemble's training trajectories. Specifically, the heat-capacity exponent $\alpha$ describes the growth rate of loss variance, $\heat = \operatorname{Var}_r\Hloss$, approaching $e_c$ from below ($t < 0$). The order-parameter exponent $\beta$ measures the onset of the mean norm $\langle S\rangle_r$ relative to $S_c = \langle S\rangle_r(e_c)$ above the transition ($t > 0$). The susceptibility exponent $\gamma$ governs the relaxation of norm variance, $\chis = \operatorname{Var}_r S$, following the transition. 
The correlation-length exponent $\nu$ controls how the transition sharpens with system size, $\Delta t\propto L^{-1/\nu}$ (Appendix~\ref{app:rgbackground}). For modular addition the label depends on the inputs only through the cyclic coordinate $(x_1+x_2)\bmod p$, whose circumference is $p$; we therefore use $p$ as the size (see Appendix~\ref{app:modulus} for the justification), and the width of the transition is measured by the relative spread over runs of $e_{50}$, the epoch at which a run first reaches 50\% test accuracy,
\begin{equation}
    \frac{\operatorname{std}_r e_{50}(p,r)}{\langle e_{50}(p,r)\rangle_r}\propto p^{-1/\nu},
    \label{eq:nu}
\end{equation}
fitted over seven moduli. The field exponent $\delta$ is measured directly from the critical isotherm of the order parameter defined in Table~\ref{tab:dictionary}. At $e = e_c$, we vary the conjugate field $h = -(\lambda - \lambda_c)/2$ and fit the singular field dependence of $m(h) = \langle S/N \rangle_r$ to $m_{\text{sing}}(h) \propto |h|^{1/\delta}$.

\paragraph{Scaling relations.}
The RG values satisfy the scaling identities by construction (Section~\ref{sec:rg}). The direct values of $\alpha$, $\beta$, $\gamma$, $\nu$, and $\delta$ come from separate fits, so for them the identities can be evaluated: Rushbrooke, $\alpha+2\beta+\gamma=2.119$ against $2$; Josephson, $d\nu=1.400$ against $2-\alpha=1.451$; Widom, $\beta(\delta-1)=0.546$ against $\gamma=0.480$. As a numerical consistency check on the dimension, the Josephson relation applied to the direct values gives $d_{\mathrm{eff}}=(2-\alpha)/\nu=1.451/1.400=1.04$, close to the one-dimensional value $d=1$ adopted for the Fiedler chain. Across the five exponents the direct and RG values differ by $1.2$--$9.7\%$.

\section{Conclusion}
\label{sec:conclusion}

We have demonstrated that the scaling and renormalization group (RG) formalism of statistical physics can be made operational for NNs, yielding an effective, quantitative description of grokking. This framework relies on three core constructions. First, a dictionary motivated by an effective Gibbs formulation of the training objective maps standard scaling quantities onto empirical NN observables, where the training epoch parameterizes an inverse temperature, weight decay serves as the field conjugate to the squared weight norm, and cross-run ensemble fluctuations serve as proxies for the susceptibility and heat capacity. Second, a spatial manifold for hidden neurons, which is intrinsically absent in neural architectures, is constructed using the exact second derivatives of the loss in per-neuron norm coordinates. Ordering neurons along the Fiedler vector of the resulting coupling graph arranges them on a one-dimensional chain where inter-neuron coupling decays monotonically with distance, unlike under random ordering. Third, blocking adjacent neurons along this chain and fitting the transformation between original and coarsened couplings around the grokking center produces a local RG map with two relevant scaling directions, from which five critical exponents follow via standard scaling relations.

The scaling exponents predicted by this local RG map agree with direct measurements on the trained ensemble within 1--10\%. Furthermore, the correlation-length exponent derived from finite-size scaling across seven task moduli, measured independently of the RG calculations, agrees with the RG prediction within 3\%. Rescaling the transition curves of seven distinct moduli by the RG exponent leads to excellent data collapse, while applying the Josephson hyperscaling relation to direct exponent estimates yields an effective dimensionality close to that of the constructed chain. Together, these findings support the RG as a quantitative tool for analyzing grokking in NNs.

\bibliography{references}

\clearpage
\appendix
\thispagestyle{empty}
\onecolumn
\aistatstitle{Scaling and Renormalization in Neural Networks: \\ An Effective Renormalization-Group Description of Grokking \\ Supplementary Materials}

\section{Notation}
\label{app:notation}

Table~\ref{tab:notation} collects the Notations.

{\small
\begin{longtable}{@{}>{\raggedright\arraybackslash}p{3.3cm}>{\raggedright\arraybackslash}p{13.2cm}@{}}
\caption{Notation.}\label{tab:notation}\\
\toprule
Symbol & Meaning \\
\midrule
\endfirsthead
\toprule
Symbol & Meaning \\
\midrule
\endhead
\bottomrule
\endfoot
\bottomrule
\endlastfoot
\multicolumn{2}{@{}l}{\emph{Task, model, and training}} \\
$p$ & modulus of the task, $p=97$ except in the modulus scan; the size variable of the finite-size analysis \\
$x_1,x_2$;\ $y$ & operands; label $y=(x_1+x_2)\bmod p$, the cyclic task coordinate \\
$x_n$,\ $y_n$ & one-hot input pair and one-hot target of training example $n$ \\
$f_\theta$,\ $\theta$ & NN function; all trainable parameters \\
$W_1$,\ $W_2$ & first- and second-layer weights, $W_1\in\mathbb{R}^{N\times 2p}$, $W_2\in\mathbb{R}^{p\times N}$ \\
$N$,\ $N_{\mathrm{alive}}$ & number of hidden neurons, $N=500$; number of unmasked neurons in a configuration \\
$n_{\mathrm{tr}}$,\ $n_y$ & number of training pairs; number of scalar targets, $n_y=n_{\mathrm{tr}}\,p$ \\
$\sigma_{ni}$,\ $\rho_n$ & activation of neuron $i$ on example $n$, $[(W_1x_n)_i]^2$; residual $W_2\sigma_n-y_n$ \\
$\lambda$ & weight decay \\
$(\beta_1,\beta_2)$ & AdamW momentum parameters \\
$e$ & training epoch \\
$r$,\ $R$ & run (seed) index; number of runs, $R=30$ \\
$\langle\cdot\rangle_r$, $\operatorname{Var}_r$, $\operatorname{std}_r$ & mean, variance, and standard deviation over runs \\
\midrule
\multicolumn{2}{@{}l}{\emph{Observables and control coordinates}} \\
$w_i$ & concatenated incoming and outgoing weights of neuron $i$ \\
$s_i$ & norm coordinate of neuron $i$, $\lVert w_i\rVert_2^2$; plays the role of a spin \\
$S$,\ $S_c$ & order parameter $\sum_i s_i$ (total squared weight norm); its mean at $\ec$, $S_c=\langle S\rangle_r(\ec)$ \\
$m(h)$,\ $m_{\mathrm{sing}}$ & order parameter per neuron at $\ec$ as a function of the field, $\langle S/N\rangle_r$; its singular part \\
$\Hloss$ & training loss (mean-squared error), the energy of the dictionary \\
$H_{\mathrm{wd}}$ & regularized objective, $\Hloss+(\lambda/2)S$ \\
$\chis$ & susceptibility proxy, $\operatorname{Var}_r S$ \\
$\heat$ & heat capacity (loss-only proxy), $\operatorname{Var}_r\Hloss$ \\
$(\ec,\lambda_c)$ & grokking center, $(145,15)$ \\
$t$ & reduced temperature, $(e-\ec)/\ec$ \\
$h$ & field conjugate to $S$, $-(\lambda-\lambda_c)/2$ \\
$\beta_{\mathrm{eff}}$,\ $t_T$ & effective inverse temperature, $\beta_{\mathrm{eff}}\propto e$; thermodynamic reduced temperature, $\ec/e-1\approx-t$ \\
\midrule
\multicolumn{2}{@{}l}{\emph{Construction of the real space}} \\
$a_i$ & gain of neuron $i$, a scalar factor on its activation \\
$g_i$,\ $G_{ij}$ & gradient and Hessian of $\Hloss$ in gain coordinates \\
$J_{ij}$ & coupling, $\partial^2\Hloss/\partial s_i\partial s_j$ \\
$A_{ij}$,\ $D$,\ $\mathcal{L}$ & adjacency $|J_{ij}|$ of the coupling graph; degree matrix; graph Laplacian $D-A$ \\
$\nabla$ & incidence matrix of the coupling graph, $\mathcal{L}=\nabla^{\top}\nabla$ \\
$\mu_k$ & eigenvalues of $\mathcal{L}$; $\mu_2$ is the algebraic connectivity \\
$v_2$ & Fiedler vector, the eigenvector of $\mu_2$ \\
$\pi(i)$,\ $\pi^{-1}(k)$ & position of neuron $i$ in the Fiedler order; neuron at position $k$ \\
$r_{ij}$ & Fiedler distance, $|\pi(i)-\pi(j)|$ \\
\midrule
\multicolumn{2}{@{}l}{\emph{Renormalization group}} \\
$b$ & scale factor of one blocking step, $b=2$ \\
$B_k$;\ $B,B'$ & block of the neurons at positions $2k-1$ and $2k$; generic blocks \\
$s'_B$,\ $J'_{BB'}$ & block spin (block mean) and block coupling \\
$K_1$,\ $K_2$ & mean $|J|$ on the first and second off-diagonal bands in Fiedler order \\
$\Kvec$,\ $\Kpvec$ & coupling vector $(K_1,K_2)$ before and after one blocking step \\
$R_b$,\ $\Kvec^*$ & RG map on coupling space, $\Kpvec=R_b(\Kvec)$; a fixed point of $R_b$ \\
$\mathcal{B}$,\ $\mathcal{A}$ & blocking map and reduced Hamiltonian in the general RG step \\
$w_n$ & Gaussian weight of point $n$ of the 65-point parameter cloud \\
$\sigma_e$,\ $\sigma_\lambda$ & widths of the Gaussian window in epoch and weight decay \\
$c$,\ $\Tmat$ & intercept and linear part (Jacobian) of the fitted affine map $\Kpvec\approx c+\Tmat\Kvec$ \\
$\Lambda_\pm$;\ $\Lambda_t$,\ $\Lambda_h$ & eigenvalues of $\Tmat$; the thermal and the magnetic one \\
$v_\pm$,\ $\ell_a$ & right and left eigenvectors of $\Tmat$ \\
$u_a$;\ $u_t$,\ $u_h$ & scaling fields, $\ell_a^{\top}\delta\Kvec$; the thermal and the magnetic one \\
$y_t$,\ $y_h$ & RG eigenvalues, $\log_2|\Lambda_t|$ and $\log_2|\Lambda_h|$ \\
$d$,\ $d_{\mathrm{eff}}$ & spatial dimension, $d=1$ for the Fiedler chain; dimension given by the Josephson relation with the direct exponents \\
$f_s$ & singular part of the free-energy density \\
$\tilde b$ & accumulated rescaling after $n$ blocking steps, $b^{n}$ \\
$\Phi$,\ $\Phi_\pm$;\ $\Delta$ & scaling functions of the free energy; gap exponent $y_h/y_t$ \\
\midrule
\multicolumn{2}{@{}l}{\emph{Critical exponents and finite size}} \\
$\alpha,\beta,\gamma,\nu,\delta$ & exponents of the specific heat, order parameter, susceptibility, correlation length, and critical isotherm \\
$\xi$ & correlation length \\
$L$,\ $\hat L$ & linear system size; the same in units of the lattice spacing \\
$\Delta t$ & width of the transition in $t$ \\
$e_{50}(p,r)$ & epoch at which run $r$ at modulus $p$ first reaches 50\% test accuracy \\
$\langle e_{50}\rangle_p$,\ $t_{50,r}$ & mean pseudo-transition epoch at modulus $p$; reduced pseudo-transition of run $r$ \\
$\nu_{\mathrm{RG}}$ & RG value of $\nu$, $1.444$ \\
$\kappa$,\ $e_\infty$ & shift exponent and limiting value of $\langle e_{50}\rangle_p$ \\
\midrule
\multicolumn{2}{@{}l}{\emph{Statistical-physics background and Gibbs form}} \\
$T$,\ $T_c$ & temperature; critical temperature \\
$M$,\ $E$,\ $C$,\ $\chi$ & magnetization, energy, specific heat, and susceptibility of a spin system \\
$\mathcal{H}$,\ $K$ & reduced Ising Hamiltonian; its nearest-neighbour coupling \\
$P(s';s)$ & block-spin projector \\
$G(r)$ & connected correlation function of two spins at distance $r$ \\
$\psi$ & finite-size scaling function \\
$\beta$ & in Appendix~\ref{app:gibbs}: inverse temperature of the Gibbs form (elsewhere: the exponent) \\
$Z$,\ $F$,\ $U$,\ $S_{\mathrm{th}}$ & partition function, free energy, internal energy, and entropy \\
$f$,\ $u$,\ $c_h$ & free energy and internal energy per neuron; specific heat at fixed field \\
$P_{\beta,h}$ & Gibbs density \\
$\eta$,\ $\mathsf{M}$ & learning rate; Hessian of the loss at a minimum \\
\end{longtable}
}

\section{Construction of the Real-Space Neuron Network}
\label{app:realspace}

\begin{figure}[t]
\centering
\includegraphics[width=\textwidth]{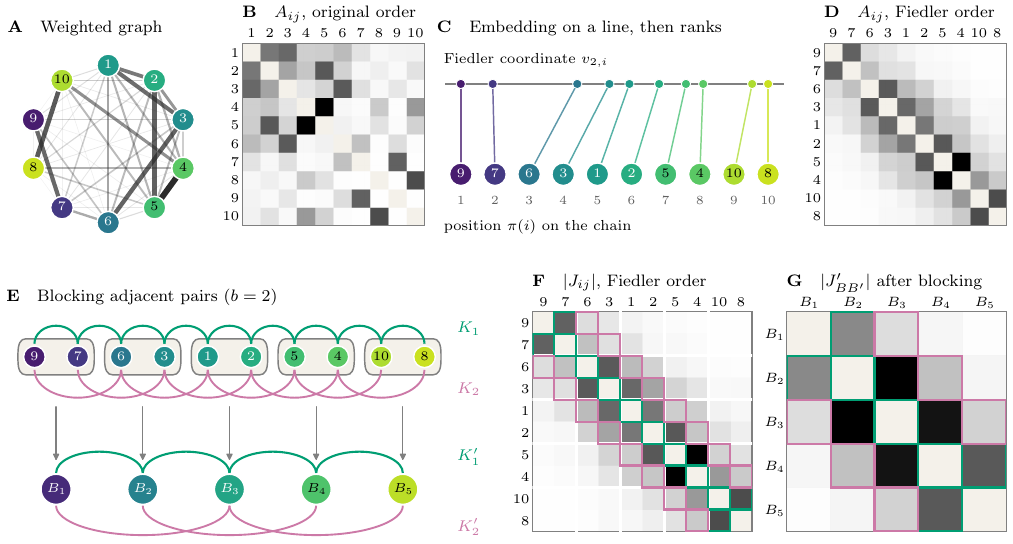}
\caption{\textbf{Steps~1--8 on a toy graph of ten vertices} (an illustration, not data from the trained NN). 
The toy weights decay with the distance between hidden positions on a line, up to multiplicative noise, and the vertex labels are shuffled. 
\textbf{(A)}~The weighted graph; the width and darkness of an edge increase with $A_{ij}$, and the colour of a vertex is its Fiedler coordinate $v_{2,i}$. 
\textbf{(B)}~The adjacency matrix in the original order (darker is larger; the diagonal is not used). 
\textbf{(C)}~The Fiedler vector places the vertices on a line with unequal spacing (top); keeping only the ranks $\pi(i)$ gives a chain of equally spaced sites (bottom). 
\textbf{(D)}~The same matrix in Fiedler order, in which the large entries lie near the diagonal. 
\textbf{(E)}~Adjacent sites of the chain are merged into the blocks $B_1,\dots,B_5$ ($b=2$); $K_1$ and $K_2$ are the mean couplings between first and second neighbours on the chain (green and pink arcs), and $K'_1$ and $K'_2$ are the same quantities on the chain of blocks. 
\textbf{(F)}~In matrix form, with $\lvert J_{ij}\rvert=A_{ij}$, the blocks are the $2\times2$ cells separated by white lines, and $K_1$ and $K_2$ are the means of the first and second off-diagonal bands (green and pink outlines). 
\textbf{(G)}~The block couplings $J'_{BB'}$ of Equation~\eqref{eq:block}, each the sum of one $2\times2$ cell, with the bands that give $K'_1$ and $K'_2$. For the NN, the corresponding matrices are shown in Figure~\ref{fig:space}.}
\label{fig:toychain}
\end{figure}

Section~\ref{sec:space} places the hidden neurons on a one-dimensional chain with the graph Laplacian. This section introduces the details following the order: a weighted graph and its adjacency matrix, the Laplacian, its spectrum, and the embedding of the graph on a line by the Fiedler vector \citep{fiedler1973algebraic}.  Figure~\ref{fig:toychain} illustrates the steps on a toy graph.

The construction is evaluated on a single NN configuration, corresponding to the state of one training run at a given epoch. This configuration is defined by the weights $W_1 \in \mathbb{R}^{N \times 2p}$ and $W_2 \in \mathbb{R}^{p \times N}$ at that epoch, alongside the $n_{\mathrm{tr}} = \lfloor p^2/2 \rfloor = 4704$ training pairs $(x_n, y_n)$ of the run. The hidden activations, residuals, and mean-squared loss are
\begin{equation}
    \sigma_{ni} = \big[(W_1 x_n)_i\big]^2, \qquad
    \rho_n = W_2 \sigma_n - y_n, \qquad
    \Hloss = \frac{1}{n_y} \sum_{n=1}^{n_{\mathrm{tr}}} \lVert \rho_n \rVert_2^2,
    \label{eq:app-loss}
\end{equation}
where $n_y = n_{\mathrm{tr}} p$ is the total number of scalar targets. Figure~\ref{fig:space} and the quantitative metrics in Section~\ref{sec:space} use the configuration at the grokking center, $(e, \lambda) = (145, 15)$. The local RG fit of Section~\ref{sec:rg} applies this construction across a parameter cloud of 65 configurations, each corresponding to a distinct $(e, \lambda)$ pair centered around this point.

\textbf{Step 1: Vertices.} \emph{Principle.} For a graph $\mathcal{G} = (\mathcal{V}, \mathcal{E})$, any real-valued scalar function defined on the vertex set $\mathcal{V}$ corresponds directly to a state vector $v \in \mathbb{R}^{|\mathcal{V}|}$.

\emph{Instantiation.} The vertices correspond to the hidden neurons. To prevent decoupled vertices resulting from weight collapse, neuron $i$ is retained only if its norm coordinate satisfies $s_i \ge 10^{-8} \operatorname{median}_j s_j$, as defined in equation~\eqref{eq:order}, with masked entries set to zero. The parameter $N_{\mathrm{alive}}$ denotes the number of surviving neurons, where all $N = 500$ neurons pass this threshold at $e_c = 145$. The one-dimensional spatial coordinate constructed below is a real-valued scalar function defined over these active vertices.

\textbf{Step 2: Edge weights.} \emph{Principle.} In a weighted graph, each edge weight quantifies the coupling strength between its incident vertices. For spatial structures such as meshes or point clouds, edge weights are determined directly from node positions, typically via a distance-dependent kernel, thus embedding the spatial metric into the graph matrix.

\emph{Instantiation.} Because hidden neurons lack physical coordinates, edge weights are instead derived from their functional coupling within the loss landscape. Treating the norm coordinates as effective spin variables and $\Hloss$ as the Hamiltonian energy, the pair interaction corresponds to the curvature $J_{ij} = \partial^2 \Hloss / \partial s_i \partial s_j$. Direct differentiation with respect to $s_i$ is ill-defined because $s_i$ aggregates multiple weight entries. To resolve this, we introduce a scalar gain $a_i$ that rescales each neuron activation as $\sigma_{ni} \to a_i \sigma_{ni}$. Because $\Hloss$ is quadratic in $a$, its derivatives at $a = \mathbf{1}$ are exact:
\begin{equation}
    g_i = \frac{\partial \Hloss}{\partial a_i} \bigg|_{a=\mathbf{1}} = \frac{2}{n_y} \sum_{n} \sigma_{ni} (\rho_n^{\top} W_2)_i, \qquad
    G_{ij} = \frac{\partial^2 \Hloss}{\partial a_i \partial a_j} \bigg|_{a=\mathbf{1}} = \frac{2}{n_y} (\sigma^{\top} \sigma)_{ij} (W_2^{\top} W_2)_{ij} .
    \label{eq:app-gain}
\end{equation}
Identifying $a_i$ with a uniform weight rescaling $w_i \to a_i w_i$ yields $s_i(a_i) = a_i^2 s_i(1)$. Under such a rescaling the output contribution of a quadratic neuron scales as $a_i^3$ rather than $a_i$, which multiplies every off-diagonal $J_{ij}$ by the constant $9$; since only ratios of couplings enter the Fiedler ordering, and $\Kvec$ and $\Kpvec$ scale together so that $\Tmat$ is unchanged, this factor drops out of all results. The chain rule at $a = \mathbf{1}$ then converts these gain-coordinate derivatives to norm coordinates:
\begin{equation}
    J_{ij} = \frac{\partial^2 \Hloss}{\partial s_i \partial s_j} = \frac{G_{ij}}{4 s_i s_j} \quad (i \neq j), \qquad
    J_{ii} = \frac{\partial^2 \Hloss}{\partial s_i^2} = \frac{G_{ii} - g_i}{4 s_i^2} .
    \label{eq:app-J}
\end{equation}
The off-diagonal entries yield equation~\eqref{eq:J} of the main text. The diagonal entries, which include the additional term $-g_i$, are omitted as the graph contains no self-loops.

\textbf{Step 3: Adjacency matrix.} \emph{Principle.} The adjacency matrix collects the edge weights, $A_{ij}\ge0$ for $i\neq j$ and $A_{ii}=0$. For an undirected graph it is real and symmetric, so its eigenvalues are real and its eigenvectors form an orthonormal basis. Sparse graphs are often obtained by keeping only the nearest neighbours of each vertex or the vertices within a fixed radius.

\emph{Instantiation.} The adjacency matrix of the kept neurons is
\begin{equation}
    A_{ij}=\tfrac12\,\lvert J_{ij}+J_{ji}\rvert\quad(i\neq j),\qquad A_{ii}=0 .
    \label{eq:app-A}
\end{equation}
The symmetrization only removes floating-point asymmetry, since the exact $J$ is symmetric, and the absolute value keeps the weights nonnegative, so that the sign of a coupling does not enter the geometry. No sparsification is applied: every pair of kept neurons is joined by an edge.

\textbf{Step 4: Graph Laplacian.} \emph{Principle.} The incidence matrix $\nabla$ maps a vertex function to its weighted differences along the edges, $(\nabla v)_{(i,j)}=\sqrt{A_{ij}}\,(v_j-v_i)$, and acts as a discrete derivative. The graph Laplacian $\mathcal{L}=\nabla^{\top}\nabla=D-A$, with the degree matrix $D_{ii}=\sum_j A_{ij}$, acts as $(\mathcal{L}v)_i=\sum_j A_{ij}(v_i-v_j)$, and its quadratic form
\begin{equation*}
    v^{\top}\mathcal{L}v=\sum_{i<j}A_{ij}\,(v_i-v_j)^2
\end{equation*}
measures how much $v$ changes across strongly weighted edges. $\mathcal{L}$ is therefore symmetric and positive semidefinite, with eigenvalues $0=\mu_1\le\mu_2\le\dots\le\mu_{N_{\mathrm{alive}}}$. The normalized variants $D^{-1/2}\mathcal{L}D^{-1/2}$ and $D^{-1}\mathcal{L}$ are also in common use.

\emph{Instantiation.} We use the unnormalized Laplacian $\mathcal{L}=D-A$ of the coupling graph, as in Section~\ref{sec:space}.

\textbf{Step 5: Connectivity and the Fiedler value.} \emph{Principle.} Every row of $\mathcal{L}$ sums to zero, so $\mathcal{L}\mathbf{1}=0$: the constant vector is an eigenvector with $\mu_1=0$. The multiplicity of the eigenvalue $0$ equals the number of connected components, so $\mu_2>0$ exactly when the graph is connected. The value $\mu_2$ is the algebraic connectivity, or Fiedler value, of the graph: the larger it is, the more strongly the graph is connected \citep{fiedler1973algebraic}. All other eigenvectors are orthogonal to $\mathbf{1}$.

\emph{Instantiation.} Since every pair of kept neurons is joined by an edge, the coupling graph is connected and $\mu_2>0$; $\mu_2$ is recorded for every configuration as a check.

\textbf{Step 6: Embedding on a line.} \emph{Principle.} To map a graph onto a line so that strongly connected vertices stay close, one minimizes $v^{\top}\mathcal{L}v$ over the coordinates $v$. The constraint $\lVert v\rVert_2=1$ fixes the scale, and $v\perp\mathbf{1}$ excludes the constant solution, which would place every vertex at the same point. By the variational characterization of the eigenvalues of a symmetric matrix, the minimizer is the eigenvector of $\mu_2$, the Fiedler vector:
\begin{equation}
    v_2 = \arg\min_{\lVert v \rVert_2 = 1,\ v \perp \mathbf{1}} \sum_{i<j} A_{ij} (v_i - v_j)^2 .
    \label{eq:app-fiedler}
\end{equation}
Adding the next few eigenvectors gives an embedding in several dimensions, the starting point of spectral clustering, while the signs of $v_2$ alone split the graph into two parts (spectral bisection).

\emph{Instantiation.} We keep only $v_2$, because the real-space RG of Section~\ref{sec:rg} blocks neighbouring sites of a one-dimensional chain. Its overall sign is arbitrary, so the ordering is defined up to a global reversal, which leaves every distance unchanged.

\textbf{Step 7: From the embedding to a lattice.} \emph{Principle.} The entries of the Fiedler vector are real coordinates whose spacing need not be uniform, whereas the sites of a lattice are equally spaced.

\emph{Instantiation.} Only the order of the entries is kept. The position of neuron $i$ is the rank of $v_{2,i}$ in ascending order, $\pi(i)\in\{1,\dots,N_{\mathrm{alive}}\}$, with ties broken by neuron index, and the distance between two neurons is $r_{ij}=\lvert\pi(i)-\pi(j)\rvert$, normalized as $r_{ij}/(N_{\mathrm{alive}}-1)$ when configurations with different $N_{\mathrm{alive}}$ are compared. The neurons thus sit on a chain of $N_{\mathrm{alive}}$ equally spaced sites (Figure~\ref{fig:toychain}C), and the result is unchanged by any monotone rescaling of $v_2$. The construction is repeated independently for every configuration, so the position of a given neuron can change from one configuration to the next; only quantities computed within a configuration, such as the coupling coordinates of Section~\ref{sec:rg}, are compared across configurations.

\textbf{Step 8: Partition into blocks.} \emph{Principle.} Spectral clustering uses the embedding to partition the vertices into a few groups, as a relaxation of cutting the graph so that the edges between groups carry little weight.

\emph{Instantiation.} The RG needs a much finer partition that follows the chain. The kept neurons are listed in Fiedler order, $\pi^{-1}(1),\pi^{-1}(2),\dots$, and grouped into consecutive pairs, the blocks $B$ of equation~\eqref{eq:block} (Figure~\ref{fig:toychain}E--G); when $N_{\mathrm{alive}}$ is odd, the last neuron is dropped. Because the ordering places more strongly coupled neurons closer together, the two neurons of a block are on average more strongly coupled than an arbitrary pair, as neighbouring spins are on a physical lattice. The block spins and block couplings are given by equation~\eqref{eq:block}, and the coordinates $(K_1,K_2)$ and $(K'_1,K'_2)$ of equation~\eqref{eq:K} are the mean absolute values of $J$ and $J'$ on their first two off-diagonal bands.

\begin{figure}[t]
    \centering
    \includegraphics[width=0.55\textwidth]{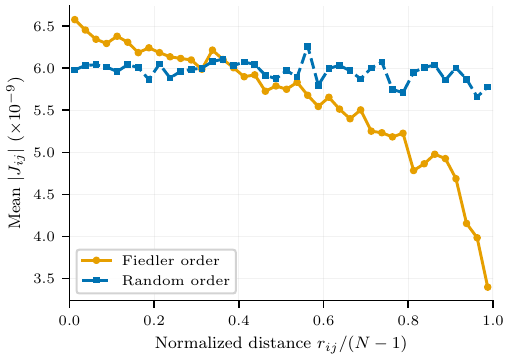}
    \caption{Distance profile at $e_c = 145$: mean $\lvert J_{ij} \rvert$ versus normalized distance across 40 equal bins for Fiedler and random orderings.}
    \label{fig:locality}
\end{figure}

\textbf{Locality diagnostics.} Two diagnostics check that the ordering produces a local geometry, in which the coupling decreases with distance. Both are shown at $\ec=145$ in Figure~\ref{fig:space}:
\begin{enumerate}
\item \emph{Block-averaged coupling.} Averaging $\lvert J_{ij}\rvert$ over $20\times20$-neuron blocks gives a diagonal band in Fiedler order that is absent in the original neuron order.
\item \emph{Distance profile.} Binning the off-diagonal pairs into 40 equal intervals of normalized distance shows that the mean $\lvert J_{ij}\rvert$ decreases with distance in Fiedler order and stays flat for a random permutation of the same matrix (Figure~\ref{fig:locality}); the ratio between the first and the last bin is the number quoted in Section~\ref{sec:space}.
\end{enumerate}

\section{Background: Real-Space Renormalization}
\label{app:rgbackground}

This section summarizes the real-space renormalization group (RG) framework, defining block-spin transformations, linearized RG flows, relevant scaling fields, and the scaling relations connecting critical exponents to RG eigenvalues \citep{wilson1975rg,cardy1996scaling}. Here and in Appendix~\ref{app:exponents}, $t=(T-T_c)/T_c$ is the thermodynamic reduced temperature, which corresponds to $t_T\approx-t$ of the main text.

\subsection{Critical Exponents and Universal Scaling}

In a standard lattice spin model defined by a reduced Hamiltonian $\mathcal{H}(s)$, the thermodynamic properties near a continuous phase transition are governed by power-law divergences in terms of reduced temperature $t = (T - T_c) / T_c$ and conjugate field $h$:
\begin{equation}
    C \propto |t|^{-\alpha}, \qquad m \propto (-t)^{\beta} \quad (t<0, h=0), \qquad \chi \propto |t|^{-\gamma}, \qquad m \propto |h|^{1/\delta} \quad (t=0), \qquad \xi \propto |t|^{-\nu} .
    \label{eq:app-exponents-def}
\end{equation}
The correlation length $\xi$ parameterizes the spatial decay of connected spin correlations $G(r) = \langle s_i s_j \rangle - \langle s_i \rangle \langle s_j \rangle \sim e^{-r/\xi}$ and constitutes the sole characteristic length scale near criticality. Under universality, system details at microscopic scales do not affect these exponents, which depend only on spatial dimensionality $d$, symmetry, and the range of the interactions.

\subsection{Block-Spin Coarse-Graining and Linearized Maps}

Real-space RG systematically coarse-grains microscopic degrees of freedom by grouping spins into blocks of volume $b^d$. Integrating out fast modes produces an effective block Hamiltonian $e^{-\mathcal{H}'(s')} = \operatorname{Tr}_s P(s';s) e^{-\mathcal{H}(s)}$, where $P(s';s)$ is a projection operator satisfying $\sum_{s'} P(s';s) = 1$. Rescaling lengths by $b$ restores the lattice spacing, transforming the correlation length via $\xi' = \xi / b$.

This transformation induces a dynamical map in coupling space $\Kvec' = R_b(\Kvec)$. Near a critical fixed point $\Kvec^* = R_b(\Kvec^*)$ with $\xi = \infty$, linearizing the map yields
\begin{equation}
    K'_a - K^*_a = \sum_{a'} T_{aa'} (K_{a'} - K^*_{a'}), \qquad T_{aa'} = \frac{\partial K'_a}{\partial K_{a'}} \bigg|_{\Kvec^*} .
    \label{eq:app-linear}
\end{equation}
The left eigenvectors of $\Tmat$ define scaling fields $u_i$ that transform multiplicatively as $u'_i = \Lambda_i u_i = b^{y_i} u_i$, where $y_i$ are the RG eigenvalues. Eigenvalues with $y_i > 0$ correspond to relevant scaling directions that expand away from the fixed point under coarse-graining.

For systems governed by two relevant scaling fields (thermal field $u_t$ with eigenvalue $b^{y_t}$ and magnetic field $u_h$ with eigenvalue $b^{y_h}$), the singular part of the free energy satisfies the homogeneous scaling relation
\begin{equation}
    f_s(u_t, u_h) = b^{-nd} f_s\big(b^{n y_t} u_t, b^{n y_h} u_h\big) .
    \label{eq:app-homog}
\end{equation}
Setting $b^{n y_t} |u_t| = O(1)$ yields $f_s(t, h) = |t|^{d/y_t} \Phi\big(h / |t|^{y_h/y_t}\big)$ for a universal scaling function $\Phi$. Differentiating $f_s$ with respect to control fields yields the standard exponent identities:
\begin{equation}
    \alpha = 2 - \frac{d}{y_t}, \qquad \beta = \frac{d - y_h}{y_t}, \qquad \gamma = \frac{2y_h - d}{y_t}, \qquad \delta = \frac{y_h}{d - y_h}, \qquad \nu = \frac{1}{y_t} .
    \label{eq:app-thermo}
\end{equation}

\subsection{Finite-Size Scaling and Task Modulus}
\label{app:modulus}

In a system of finite linear spatial dimension $L$, the inverse scale $\hat{L}^{-1} = a / L$ transforms under blocking as $\hat{L}^{-1} \to b \hat{L}^{-1}$, behaving as a relevant scaling parameter with eigenvalue $y_L = 1$. The susceptibility then follows the finite-size scaling hypothesis:
\begin{equation}
    \chi(t, \hat{L}^{-1}) \simeq \hat{L}^{\gamma/\nu} \psi\big(t \hat{L}^{1/\nu}\big) .
    \label{eq:app-fss}
\end{equation}

In our neural network formulation, the finite-size parameter is provided by the task modulus $p$ rather than the neuron count $N$. For modular addition $y = (x_1 + x_2) \bmod p$, the cyclic representation on $\mathbb{Z}_p$ introduces a fundamental infrared wave-number cutoff $q_{\min} = 2\pi / p$. 

This mapping rests on two primary hypotheses:
\begin{enumerate}
    \item[(i)] Phase transitions across varying moduli $p$ are controlled by the same effective scaling regime as the local neuron map, with $p$ serving as the characteristic linear system size.
    \item[(ii)] Empirical run-to-run variances in transition epochs are dictated by fluctuations in this scaling variable, so that $z_r = p^{y_t} t_{50,r}$ follows a $p$-independent distribution.
\end{enumerate}

\begin{figure}[t]
    \centering
    \includegraphics[width=\textwidth]{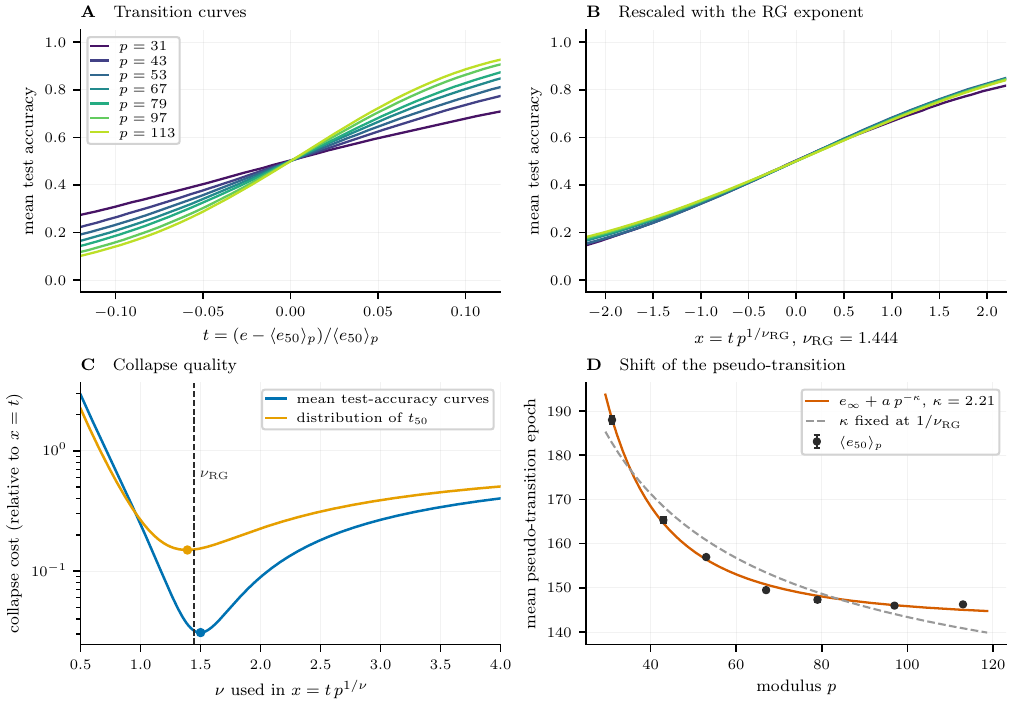}
    \caption{Finite-size collapse of the modulus scan across 350 confirmation runs ($N=500$). (A) Mean test accuracy against reduced epoch $t$, centered at mean pseudo-transition $\langle e_{50} \rangle_p$. (B) Collapse achieved against rescaled coordinate $x = t p^{1/\nu_{\mathrm{RG}}}$. (C) Collapse cost function versus candidate scaling exponents, highlighting the minimum near $\nu_{\mathrm{RG}}$. (D) Scaling of mean pseudo-transition epoch against $p$, comparing unconstrained power-law fitting with a fixed $1/\nu_{\mathrm{RG}}$ decay exponent.}
    \label{fig:modcollapse}
\end{figure}

Under these assumptions, the pseudo-transition width scales as $\Delta t \propto p^{-1/\nu}$. Figure~\ref{fig:modcollapse} supports these hypotheses across 350 validation runs. Rescaling test-accuracy trajectories by $p^{1/\nu_{\mathrm{RG}}}$ with $\nu_{\mathrm{RG}} = 1.444$ produces data collapse onto a common curve, with the collapse cost minimized at $\nu = 1.50$ for mean accuracy curves and $\nu = 1.39$ for $t_{50}$ distributions.

\subsection{Dictionary Mapping to Neural Networks}

To apply this formalism to trained neural networks, physical variables are mapped directly to network observables:
\begin{enumerate}
    \item \emph{Lattice and spins.} Hidden neurons represent 1D lattice sites ($d=1$) ordered via the Fiedler vector, with norm coordinates $s_i$ acting as spin variables.
    \item \emph{Coarse-graining operator.} Spatial blocking merges adjacent Fiedler pairs ($b=2$) by taking block means of the spins and sums of the couplings.
    \item \emph{Control parameters.} Training epoch acts as an effective inverse temperature, measured by $t$, while weight decay provides the conjugate field $h$.
    \item \emph{Renormalization map.} Linearizing the empirical transformation between fine and coarsened couplings gives the $2 \times 2$ matrix $\Tmat$, whose eigenvalues yield $y_t$ and $y_h$. Exponents then follow via equation~\eqref{eq:app-thermo}.
\end{enumerate}
\section{Derivation of Critical Exponent Relations}
\label{app:exponents}

Section~\ref{sec:rg} utilizes scaling relations derived by evaluating the singular free energy $f_s$ under finite spatial rescaling \citep{cardy1996scaling}. This appendix presents the detailed line-by-line derivation of these relations given control variables $(t, h)$, spatial dimension $d$, and relevant RG eigenvalues $y_t, y_h > 0$. The exponents are defined through standard thermodynamic limit divergences near criticality: $C \propto |t|^{-\alpha}$, $m \propto (-t)^\beta$, $\chi \propto |t|^{-\gamma}$, $m \propto |h|^{1/\delta}$ at $t=0$, and $\xi \propto |t|^{-\nu}$.

\subsection{From One Blocking Step to a Free Scale Factor}

The fundamental single-step RG homogeneity relation describes how coarse-graining by a block spatial factor $b$ reduces the number of degrees of freedom by $b^d$ while inflating the scaling fields by $b^{y_t}$ and $b^{y_h}$. Iterating this coarse-graining step $n$ times yields
\begin{equation}
    f_s(t,h) = b^{-nd} f_s\big(b^{ny_t}t, \, b^{ny_h}h\big) \quad \Longrightarrow \quad f_s(t,h) = \tilde{b}^{-d} f_s\big(\tilde{b}^{y_t}t, \, \tilde{b}^{y_h}h\big), \qquad \tilde{b} \equiv b^n .
    \label{eq:exp-ell}
\end{equation}
Because this relation holds for any integer $n$, the accumulated scale factor $\tilde{b}$ is a free parameter. Selecting a specific trajectory corresponds to choosing the total spatial rescaling such that the rescaled thermal argument leaves the critical manifold. Setting the argument to unit magnitude gives
\begin{equation}
    \tilde{b}^{y_t}|t| = 1 \quad \Longleftrightarrow \quad \tilde{b} = |t|^{-1/y_t} .
    \label{eq:exp-choice}
\end{equation}
Substituting this scale factor back into equation~\eqref{eq:exp-ell} isolates the non-analytic dependence on $t$:
\begin{equation}
    f_s(t,h) = |t|^{d/y_t} f_s\big(\pm1, \, h\,|t|^{-y_h/y_t}\big) \equiv |t|^{d/y_t} \Phi_\pm\big(h / |t|^{\Delta}\big), \qquad \Delta \equiv \frac{y_h}{y_t} ,
    \label{eq:exp-scalingform}
\end{equation}
where the sign corresponds to $t > 0$ or $t < 0$, and $\Phi_\pm(x) \equiv f_s(\pm1, x)$ represents smooth scaling functions of a single variable above and below $T_c$.

\subsection{Thermodynamic Exponents}

Each thermodynamic exponent is obtained directly by differentiating equation~\eqref{eq:exp-scalingform} with respect to physical fields and extracting the resulting power laws.

\paragraph{Specific heat, $\alpha$.}
Evaluating $f_s(t,0) = \Phi_\pm(0) |t|^{d/y_t}$ at zero field and taking two derivatives with respect to the reduced temperature $t$ yields
\begin{equation}
    C \propto -\frac{\partial^2 f_s}{\partial t^2}\bigg|_{h=0} \propto |t|^{d/y_t - 2} \equiv |t|^{-\alpha} \quad \Longrightarrow \quad \alpha = 2 - \frac{d}{y_t} .
\end{equation}

\paragraph{Order parameter, $\beta$.}
Differentiating once with respect to the magnetic field $h$ brings down a factor of $|t|^{-\Delta}$ from the scaling argument:
\begin{equation}
    m = -\frac{\partial f_s}{\partial h} = -|t|^{d/y_t - \Delta} \Phi'_\pm\big(h / |t|^\Delta\big) = -|t|^{(d-y_h)/y_t} \Phi'_\pm\big(h / |t|^\Delta\big) .
    \label{eq:exp-m}
\end{equation}
On the ordered side ($t < 0$) at zero magnetic field, the spontaneous magnetization evaluates to $m \propto (-t)^{(d-y_h)/y_t} \equiv (-t)^\beta$, giving
\begin{equation}
    \beta = \frac{d - y_h}{y_t} .
\end{equation}
On the disordered side ($t > 0$), symmetry dictates $\Phi'_+(0) = 0$, so the spontaneous magnetization vanishes.

\paragraph{Susceptibility, $\gamma$.}
Taking a second derivative with respect to $h$ brings down an additional factor of $|t|^{-\Delta}$:
\begin{equation}
    \chi = \frac{\partial m}{\partial h}\bigg|_{h=0} = -|t|^{(d-2y_h)/y_t} \Phi''_\pm(0) \propto |t|^{-(2y_h - d)/y_t} \equiv |t|^{-\gamma} \quad \Longrightarrow \quad \gamma = \frac{2y_h - d}{y_t} .
\end{equation}

\paragraph{Critical isotherm, $\delta$.}
At $t=0$, the thermal scale choice in equation~\eqref{eq:exp-choice} becomes singular. We instead choose $\tilde{b} = |h|^{-1/y_h}$ to clear the field argument, transforming equation~\eqref{eq:exp-ell} into $f_s(0,h) = |h|^{d/y_h} f_s(0, \pm1)$. Differentiating with respect to field yields
\begin{equation}
    m = -\frac{\partial f_s}{\partial h}\bigg|_{t=0} \propto |h|^{d/y_h - 1} = |h|^{(d-y_h)/y_h} \equiv |h|^{1/\delta} \quad \Longrightarrow \quad \delta = \frac{y_h}{d - y_h} .
\end{equation}
Equivalently, requiring $m$ in equation~\eqref{eq:exp-m} to remain finite as $t \to 0$ at fixed $h$ forces $\Phi'_\pm(x) \sim x^{(d-y_h)/y_h}$ for large $x$, which reproduces the same exponent value.

\subsection{Correlation Length}

\paragraph{Correlation length, $\nu$.}
Under spatial coarse-graining, physical length scales shrink by the total rescaling factor $\tilde{b}$, establishing the homogeneity relation $\xi(t,h) = \tilde{b} \, \xi(\tilde{b}^{y_t} t, \, \tilde{b}^{y_h} h)$. Choosing $\tilde{b} = |t|^{-1/y_t}$ at zero magnetic field yields
\begin{equation}
    \xi(t) = |t|^{-1/y_t} \xi(\pm1, 0) \propto |t|^{-\nu} \quad \Longrightarrow \quad \nu = \frac{1}{y_t} .
\end{equation}
The choice $\tilde{b} = |t|^{-1/y_t}$ corresponds physically to iterating the RG coarse-graining process until the effective lattice block size expands to equal the underlying spatial correlation length $\xi$.

\subsection{Scaling Identities}

Expressing the five thermodynamic and correlation exponents in terms of $(y_t, y_h, d)$ leads directly to the fundamental scaling and hyperscaling identities:
\begin{equation}
\begin{aligned}
    \alpha + 2\beta + \gamma &= \Big(2 - \frac{d}{y_t}\Big) + \frac{2(d-y_h)}{y_t} + \frac{2y_h - d}{y_t} = 2 && \text{(Rushbrooke identity)}, \\
    \alpha + \beta(1+\delta) &= \Big(2 - \frac{d}{y_t}\Big) + \frac{d-y_h}{y_t} \cdot \frac{d}{d-y_h} = 2 && \text{(Griffiths identity)}, \\
    \beta(\delta - 1) &= \frac{d-y_h}{y_t} \cdot \frac{2y_h - d}{d-y_h} = \gamma && \text{(Widom identity)}, \\
    d\nu &= \frac{d}{y_t} = 2 - \alpha && \text{(Hyperscaling relation)} .
\end{aligned}
\label{eq:exp-identities}
\end{equation}
The first three identities arise purely from the properties of the free energy, whereas hyperscaling explicitly involves the spatial dimension $d$ and relies on the assumption of short-range microscopic interactions.

\subsection{Numerical Values}

Substituting $d=1$ and the unrounded empirical RG eigenvalues $(y_t, y_h) = (0.6923, 0.6515)$ into these relations yields the theoretical values used in Section~\ref{sec:rg}:
\begin{equation}
\begin{aligned}
    \alpha &= 2 - \frac{1}{0.6923} = 0.556, & \beta &= \frac{1-0.6515}{0.6923} = 0.503, & \gamma &= \frac{2(0.6515)-1}{0.6923} = 0.438, \\
    \nu &= \frac{1}{0.6923} = 1.444, & \delta &= \frac{0.6515}{1-0.6515} = 1.870 .
\end{aligned}
\end{equation}
All numerical evaluations are performed using full-precision empirical inputs prior to rounding, ensuring internal mathematical consistency across all reported quantities.

\section{Local Regression Details}
\label{app:fit}

\begin{figure}[t]
    \centering
    \includegraphics[width=\textwidth]{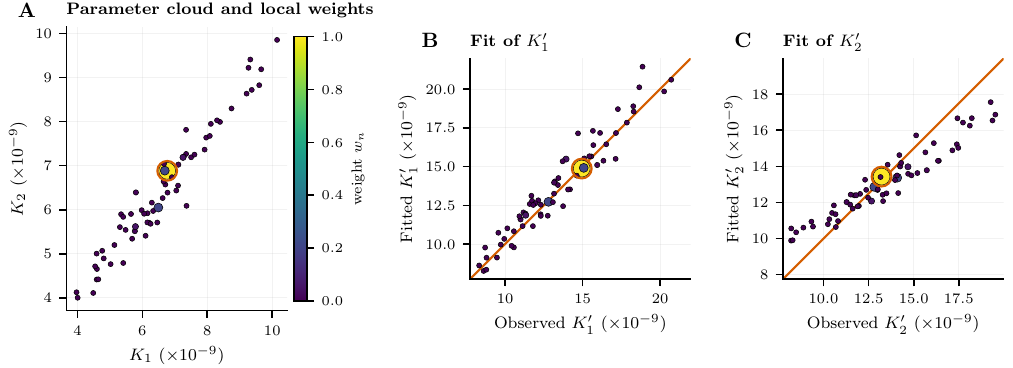}
    \caption{The regression behind the local RG map $\Tmat$ of equation~\eqref{eq:T}. (A) The 65 coupling coordinates $\Kvec_n=(K_1,K_2)_n$ of the parameter cloud (13 epochs $\times$ 5 weight decays), coloured and sized by the Gaussian weight $w_n$ of equation~\eqref{eq:fit}; the ring marks the grokking center $(\ec,\lambda_c)=(145,15)$, which carries weight 1. (B, C) Observed block coordinates $K'_1$ and $K'_2$ against the values $c+\Tmat\Kvec_n$ returned by the weighted affine fit, with the identity line. The fit is local: points far from the center have negligible weight and are shown only to indicate the extent of the cloud.}
    \label{fig:localfit}
\end{figure}

For a cloud point $n$ with epoch $e_n$ and weight decay $\lambda_n$, the weighting function in equation~\eqref{eq:fit} evaluates to
\begin{equation}
    w_n = \exp\!\left( -\frac{(e_n - e_c)^2}{2\sigma_e^2} - \frac{(\lambda_n - \lambda_c)^2}{2\sigma_\lambda^2} \right) ,
    \label{eq:cloud-weights}
\end{equation}
with the evaluation center $(e_c, \lambda_c) = (145, 15)$ and bandwidth parameters $(\sigma_e, \sigma_\lambda) = (2.833, 0.410)$. Under this weighting profile, the evaluation center receives unit weight $w=1$, its epoch neighbors at $e \in \{140, 150\}$ receive weight $0.21$, and its weight-decay neighbors at $\lambda \in \{14, 16\}$ receive weight $0.05$. Coordinates are normalized by a common scale factor prior to fitting, and a ridge penalty of $10^{-5} \lVert \Tmat \rVert_F^2$ is included for numerical stability, resulting in a matrix condition number of $208$. Figure~\ref{fig:localfit} illustrates the weighted point cloud along with the observed versus fitted values for both matrix components.

\paragraph{Labeling the eigenvectors.}
The control tangents are estimated via an unweighted linear regression of the 65 cloud coordinates $\Kvec_n$ onto the reduced epoch $(e_n - e_c) / e_c$ and the weight-decay offset $\lambda_n - \lambda_c$. This regression yields the two principal coefficient vectors, $\partial \Kvec / \partial e$ and $\partial \Kvec / \partial \lambda$.

\begin{figure}[t]
    \centering
    \includegraphics[width=0.55\textwidth]{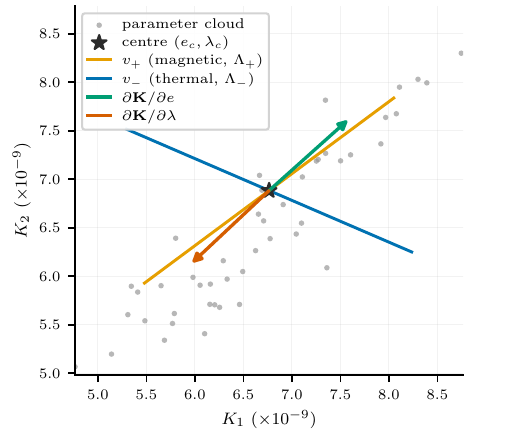}
    \caption{Eigendirections of $\Tmat$ and control tangents on the coupling plane. Grey points: the 65-point parameter cloud; star: the grokking center $(\ec,\lambda_c)=(145,15)$. The orange and blue lines are the right eigendirections $v_+$ and $v_-$ through the center; the arrows are the unit tangents $\partial\Kvec/\partial e$ and $\partial\Kvec/\partial\lambda$ obtained by regressing the cloud on the two control coordinates. }
    \label{fig:alignment}
\end{figure}

\FloatBarrier

\section{Gibbs Form of the Dictionary and the Temperature-Like Coordinate}
\label{app:gibbs}

\begin{table}[!ht]
    \centering
    \small
    \caption{Dictionary between the standard scaling analysis and the trained NN. Every entry in the right column is an operational definition, not an identity; the assumptions behind each are discussed in the text.}
    \label{tab:dictionary}
    \begin{tabular}{p{3.0cm}p{5.2cm}p{7.6cm}}
        \toprule
        Quantity & Statistical physics & NN (this work) \\
        \midrule
        \multicolumn{3}{l}{\emph{Thermodynamic quantities}} \\
        Reduced temperature $t$ & $(T-T_c)/T_c$ & $(e-\ec)/\ec$, $\ec=145$ (epoch as an inverse-temperature-like coordinate, so that $t\approx-t_T$) \\
        Field $h$ & $H=\cdots-h\sum_i s_i$ & $h=-(\lambda-15)/2$ (weight decay, conjugate to $S$) \\
        Magnetization $M$ & $\sum_i s_i$, $m=M/N$ & $S=\sum_i\lVert w_i\rVert_2^2$, measured relative to $S_c$ \\
        Susceptibility $\chi$ & $\partial m/\partial h\propto\operatorname{Var}(M)$ & $\chis=\operatorname{Var}_r[S]$ over 30 seeds (proxy) \\
        Specific heat $C$ & $-T\,\partial^2 f/\partial T^2\propto\operatorname{Var}(E)$ & $\heat=\operatorname{Var}_r[\Hloss]$ over 30 seeds (proxy) \\
        Ensemble average & Gibbs ensemble & 30 training seeds (initialization and data split) \\
        \midrule
        \multicolumn{3}{l}{\emph{Renormalization-group quantities}} \\
        Spin $s_i$ & spin on a lattice site & norm coordinate $s_i$ of neuron $i$ \\
        Couplings $\Kvec$ & Hamiltonian parameters & $J_{ij}=\partial^2\Hloss/\partial s_i\partial s_j$; $K_1,K_2$ = mean $|J|$ on the first and second neighbour bands \\
        Geometry, distance $r$ & lattice sites and their distance & graph of $|J|$, Fiedler ordering $\pi$, $r_{ij}=|\pi(i)-\pi(j)|$ \\
        Block spin, $b$ & merge $b$ adjacent spins & merge adjacent neuron pairs in Fiedler order, $b=2$; block mean \\
        Linearization $\Tmat$ & $\partial K'/\partial K$  & local regression of $\Kvec\to\Kpvec$  \\
        System size $L$ & finite-size scaling & circumference $p$ of the task coordinate $(x_1+x_2)\bmod p$ \\
        \bottomrule
    \end{tabular}
\end{table}

The dictionary in Table~\ref{tab:dictionary} is motivated by a Gibbs description that the neural network does not explicitly instantiate. Writing the training loss as $\Hloss(\theta)$ and the total squared weight norm as $S(\theta)$, the objective function with weight decay expanded around the reference value $\lambda_c$ reads
\begin{equation}
    H_{\mathrm{wd}}(\theta) = \Hloss(\theta) + \frac{\lambda_c}{2} S(\theta) - h S(\theta) , \qquad h \equiv -\frac{\lambda - \lambda_c}{2} .
    \label{eq:gibbs-wd}
\end{equation}
Here, the perturbation term $-h S$ has the form of a magnetic coupling $-h M$, where increasing $h$ relaxes the weight penalty and favors states with larger weight norms. Given a fixed reference measure $d\mu$, the partition function $Z(\beta, h) = \int e^{-\beta H_{\mathrm{wd}}(\theta)} \, d\mu(\theta)$ defines the Gibbs density $P_{\beta,h} = Z^{-1} e^{-\beta H_{\mathrm{wd}}}$. The corresponding thermodynamic potentials include the free energy $F = -\beta^{-1} \log Z$, internal energy $U = \langle H_{\mathrm{wd}} \rangle = -\partial_\beta \log Z$, and thermal entropy $S_{\mathrm{th}} = \log Z + \beta U$, satisfying $F = U - T S_{\mathrm{th}}$ with $k_B \equiv 1$. Expressed on a per-neuron basis ($f \equiv F/N$ and $u \equiv U/N$), the standard response functions are
\begin{equation}
    m = \frac{\langle S \rangle}{N} = -\left( \frac{\partial f}{\partial h} \right)_T , \qquad
    \chi = \left( \frac{\partial m}{\partial h} \right)_T = \frac{\beta}{N} \operatorname{Var}_P(S) , \qquad
    c_h = \left( \frac{\partial u}{\partial T} \right)_h = -T \left( \frac{\partial^2 f}{\partial T^2} \right)_h = \frac{\beta^2}{N} \operatorname{Var}_P(H_{\mathrm{wd}}) .
    \label{eq:thermo-responses}
\end{equation}
Because $h = -(\lambda - \lambda_c)/2$, increasing the regularization scale strength directly yields $(\partial m / \partial \lambda)_T = -\chi / 2$. 

The empirical observables in Section~\ref{sec:setup} adapt these statistical-mechanical identities via three precise substitutions. First, the formal Gibbs expectation value is replaced by an ensemble average over independent training seeds. Second, the dimensional prefactors $\beta/N$ and $\beta^2/N$ are omitted, as they leave the asymptotic power-law exponents invariant. Third, $\operatorname{Var}(H_{\mathrm{wd}})$ is replaced by the loss-only variance $\operatorname{Var}(\Hloss)$, because the full-Hamiltonian variance contains the terms $(\lambda^2/4)\operatorname{Var}(S)$ and $\lambda\operatorname{Cov}(\Hloss,S)$. While the formal specific heat involves the complete Hamiltonian, the loss variance serves as an operational surrogate in the critical regime.

The identification of training epoch with an effective inverse temperature follows the dynamical framing of \citet{zhang2023dynamical}. Near a local minimum $\theta_*$ of the loss function, defining $x \equiv \theta - \theta_*$ and the Hessian $\mathsf{M} \equiv \nabla^2 \Hloss(\theta_*)$, gradient descent with learning rate $\eta$ updates as $x_{k+1} \simeq (I - \eta \mathsf{M}) x_k$ to linear order. In the continuum limit with continuous time $\tau \equiv \eta k$, this recursion reduces to $\partial_\tau x = -\mathsf{M} x$, yielding the solution $x(\tau) = e^{-\tau \mathsf{M}} x(0)$, which formally resembles a Schr\"odinger evolution in imaginary time governed by Hamiltonian $\mathsf{M}$. Under a fixed training schedule, $\tau$ tracks the cumulative compute, motivating the identification of epoch with an effective inverse temperature $\beta_{\mathrm{eff}} \propto e$. This identification yields the reduced thermal control variable $t_T = e_c/e - 1 = -t/(1+t)$ used in Section~\ref{sec:setup}. Two key caveats qualify this effective mapping: trained models do not strictly constitute equilibrium samples from $P_{\beta,h}$, and the decoupled weight decay mechanism in AdamW does not literally reduce to adding an explicit term $-h S$ directly into the loss function.

\end{document}